%% file: example_paper.tex
\documentclass{article}

\usepackage{microtype}
\usepackage{graphicx}
\usepackage{subfigure}
\usepackage{booktabs} % for professional tables
\usepackage{csquotes}
\usepackage{tabularx}
\usepackage{natbib}
\usepackage{placeins}
\DeclareUnicodeCharacter{2060}{\nolinebreak}

\usepackage[accepted]{icml2024}
\usepackage{hyperref}
\usepackage{amsmath}
\usepackage{amssymb}
\usepackage{mathtools}
\usepackage{amsthm}

\usepackage[capitalize,noabbrev]{cleveref}

\theoremstyle{plain}
\newtheorem{theorem}{Theorem}[section]

\theoremstyle{definition}

\newtheorem{assumption}[theorem]{Assumption}
\theoremstyle{remark}

\usepackage[textsize=tiny]{todonotes}

\usepackage{tablefootnote}
\usepackage{multicol}
\usepackage{multirow}
\usepackage{booktabs}
\usepackage{verbatim}
\usepackage{kotex}
\usepackage{xcolor} 
\usepackage{xspace}
\usepackage{colortbl}
\definecolor{baselinecolor}{gray}{.9}
\newcommand{\baseline}[1]{\cellcolor{baselinecolor}{#1}}

\usepackage[labelformat=brace]{subcaption}
\usepackage{fancyhdr}
\usepackage{graphicx}

\newcommand{\MSP}{\textit{MSP}\xspace}
\newcolumntype{C}{>{$\boldmath$}c}
\usepackage{pifont}
\usepackage{enumitem}
\usepackage{algorithm}
\usepackage{algorithmic}
\usepackage{arydshln} 
\usepackage{times}
\usepackage{latexsym}
\usepackage{booktabs}
\usepackage{amssymb}
\usepackage{amsmath}
\usepackage{fixltx2e}
\usepackage[most]{tcolorbox} % For creating colored boxes
\usepackage{listings} % For code listing
\usepackage{listings}
\usepackage{mdframed} % For creating framed boxes
\definecolor{verylightgray}{gray}{0.97}

\usepackage{cuted,tcolorbox,lipsum}
\usepackage{caption}
\usepackage{capt-of}
\newtcolorbox{databox}[1][]{
  colback=verylightgray, % Very light gray background for content
  colframe=lightgray, % Lighter gray frame
  title= NLI Dataset, % Box title
  fonttitle=\small\bfseries,
  width=\linewidth,% Smaller bold title font
  #1,
}

\definecolor{myr}{RGB}{255, 77, 106}
\definecolor{myg}{RGB}{23, 168, 160}
\definecolor{myo}{RGB}{252, 163, 98}
\definecolor{light-blue}{RGB}{200, 233, 255}
\definecolor{pink}{RGB}{252, 163, 98}
\input std_macros

\definecolor{backgroundcolor}{rgb}{0.97, 0.97, 0.97}
\definecolor{commentcolor}{rgb}{0.12, 0.38, 0.18}
\definecolor{stringcolor}{rgb}{0.2, 0.6, 0.4}
\definecolor{keywordcolor}{rgb}{0.88, 0.4, 0.4}
\definecolor{numb}{rgb}{0.5,0,0.5}

\lstdefinelanguage{python}{
    basicstyle=\normalfont\ttfamily,
    numberstyle=\scriptsize,
    showstringspaces=false,
    breaklines=true,
    frame=single,
    rulecolor=\color{darkgray}, % Frame color
    backgroundcolor=\color{backgroundcolor},
    tabsize=2,
    morecomment=[s][\color{commentcolor}]{"""}{"""},
    morecomment=[l][\color{commentcolor}]{\#},
    morestring=[b][\color{stringcolor}]{"},
    morestring=[b][\color{stringcolor}]{'},
    morekeywords={as, assert, break, class, continue, def, del, elif, else, except, exec, finally, for, from, global, if, import, in, is, lambda, nonlocal, not, or, pass, raise, return, try, while, with, yield},
    keywordstyle=\color{keywordcolor}\bfseries,
    numbers=left,
    numberstyle=\tiny\color{numb},
    identifierstyle=\color{blue},
    stringstyle=\color{stringcolor},
}
\usepackage{bm}
\usepackage{url}

\usepackage{fixltx2e}
\icmltitlerunning{\textsc{KPC-cF:} Aspect-Based Sentiment Analysis via Implicit-Feature Alignment with Corpus Filtering}

\begin{document}
\twocolumn[
\icmltitle{\textsc{KPC-cF:} Aspect-Based Sentiment Analysis via\\
Implicit-Feature Alignment with Corpus Filtering}




\begin{icmlauthorlist}
  
\icmlauthor{ Jason K. Nam}{yyy,comp}
%\icmlauthor{Firstname2 Lastname2}{equal,yyy,comp}
%\icmlauthor{Firstname3 Lastname3}{comp}
%\icmlauthor{Firstname4 Lastname4}{sch}
%\icmlauthor{Firstname5 Lastname5}{yyy}
%\icmlauthor{Firstname6 Lastname6}{sch,yyy,comp}
%\icmlauthor{Firstname7 Lastname7}{comp}
%\icmlauthor{Firstname8 Lastname8}{sch}
%\icmlauthor{Firstname8 Lastname8}{yyy,comp}
\end{icmlauthorlist}

\icmlaffiliation{yyy}{Department of Data Science, Hongik University}
%\icmlaffiliation{yyy}{Department of XXX, University of YYY, Location, Country}
\icmlaffiliation{comp}{KorABSA Laboratory, MODULABS}%Company Name, Location, Country}
%\icmlaffiliation{sch}{School of ZZZ, Institute of WWW, Location, Country}

\icmlcorrespondingauthor{Kibeom Nam}{skarlqja68@gmail.com}
%\icmlcorrespondingauthor{Firstname2 Lastname2}{first2.last2@www.uk}


%\renewcommand\thefootnote{*}
%\footnotetext{Blinded}
%\footnotetext{This work was done while at AI Research, MODULABS.}
%\renewcommand\thefootnote{\arabic{footnote}}

\icmlkeywords{Machine Learning, ICML}

\vskip 0.3in
]



\printAffiliationsAndNotice{}  % leave blank if no need to mention equal contribution

\begin{abstract}
Investigations into Aspect-Based Sentiment Analysis (ABSA) for Korean industrial reviews are notably lacking in the existing literature. 
Our research proposes an intuitive and effective framework for ABSA in low-resource languages such as Korean.
It optimizes prediction labels by integrating translated benchmark and unlabeled Korean data. Using a model fine-tuned on translated data, we pseudo-labeled the actual Korean NLI set. Subsequently, we applied LaBSE and \MSP{}-based filtering to this pseudo-NLI set as implicit feature, enhancing Aspect Category Detection and Polarity determination through additional training. Incorporating dual filtering, this model bridged dataset gaps and facilitates feature alignment with minimal resources.
By implementing alignment pipelines, our approach aims to leverage high-resource datasets to develop reliable predictive and refined models within corporate or individual communities in low-resource language countries.
Compared to English ABSA, our framework showed an approximately 3\% difference in F1 scores and accuracy. %We will show the model and data for Korean ABSA, publicly available at the repository. 
%We will show the model and data for Korean ABSA, publicly available at \url{https://github.com/namkibeom/KPC-cF}.
We will release our dataset and code for Korean ABSA, at this link\footnote{\url{https://anonymous.4open.science/r/KPC-cF-21E8}}.
%\footnote{\url{https://github.com/namkibeom/KPC-cF}}.
\end{abstract}

\section{Introduction}
\label{submission}
In low-resource downstream tasks such as Korean ABSA, constraints exist in constructing ABSA systems that are socially and industrially beneficial (e.g., obtaining accurate labels and high-quality training data, building a efficient serving model). Addressing this challenge is fundamentally crucial for the practical implementation of multilingual ABSA leveraging the advantages of language models~\cite{zhang-etal-2021-cross, lin2023cl}. On the other hand, ABSA utilizing Large Language Models like ChatGPT can perform labeling through prompt tuning. 
However, due to reduced sensitivity to gradient-based tuning, the model may exhibit overconfidence in random labels, and limitations persist regarding resources for training and inference \cite{wang2023chatgpt, wu2023brief, kossen2024context}. 

\begin{figure}[t]
\centerline{\includegraphics[width=0.9\linewidth]{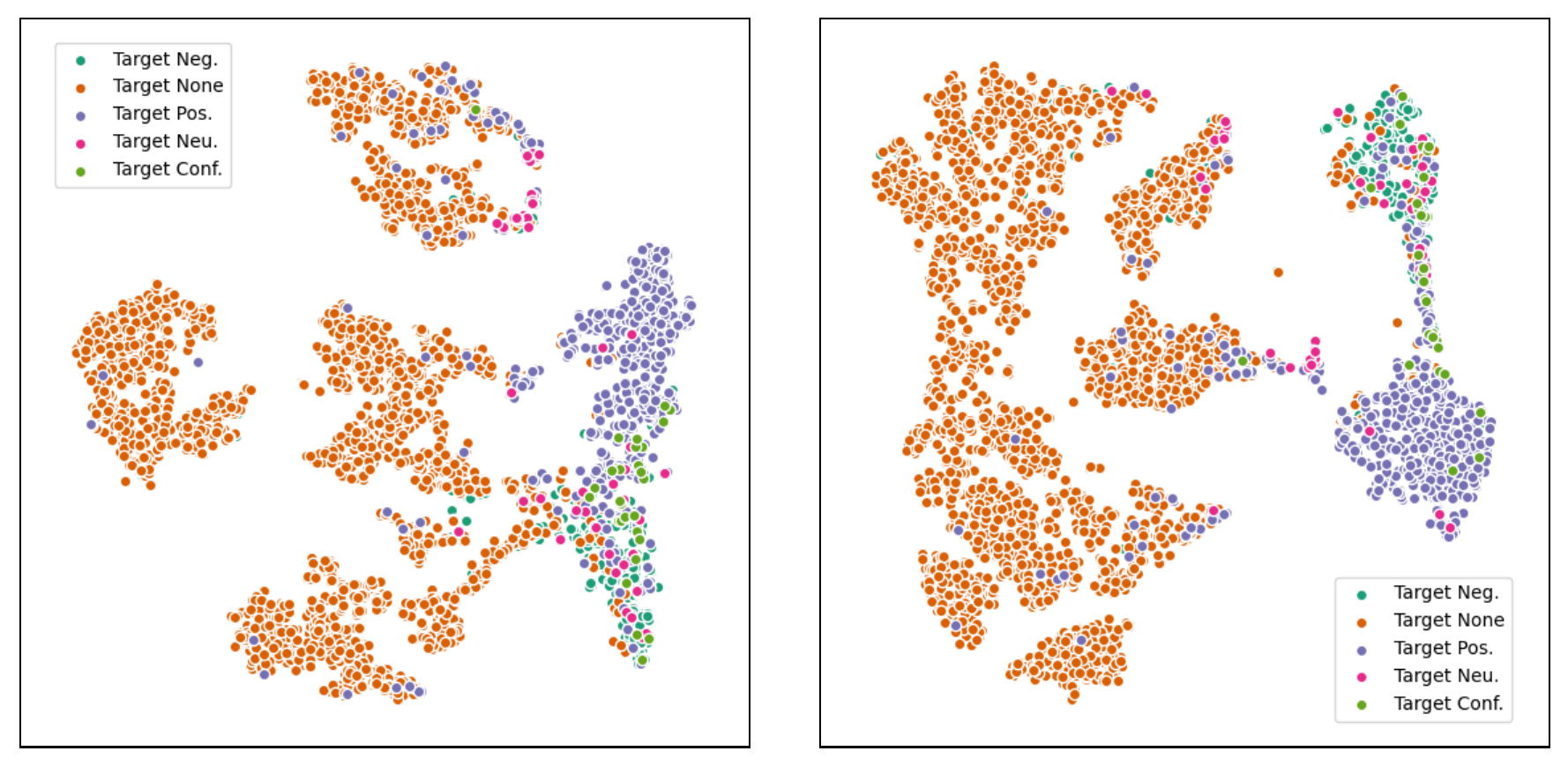}}
\footnotesize\qquad\quad\,  \textbf{(a) TR w/o PL-\textsc{cF} \qquad\qquad (b) TR w/ PL-\textsc{cF}}
\caption{t-SNE visualization of last \texttt{[CLS]} embeddings extracted from KR3 test set by two different experimental Baseline\textsubscript{XLM-R} encoders. Our \( L_\text{align} \) on filtering data encourages the encoder to produce discriminable representations of different sentiment polarities, focusing on relevant aspects.}
\label{fig3}
\end{figure}

\begin{figure*}[t]
\centerline{\includegraphics[width=0.975\textwidth, height=7.75cm]{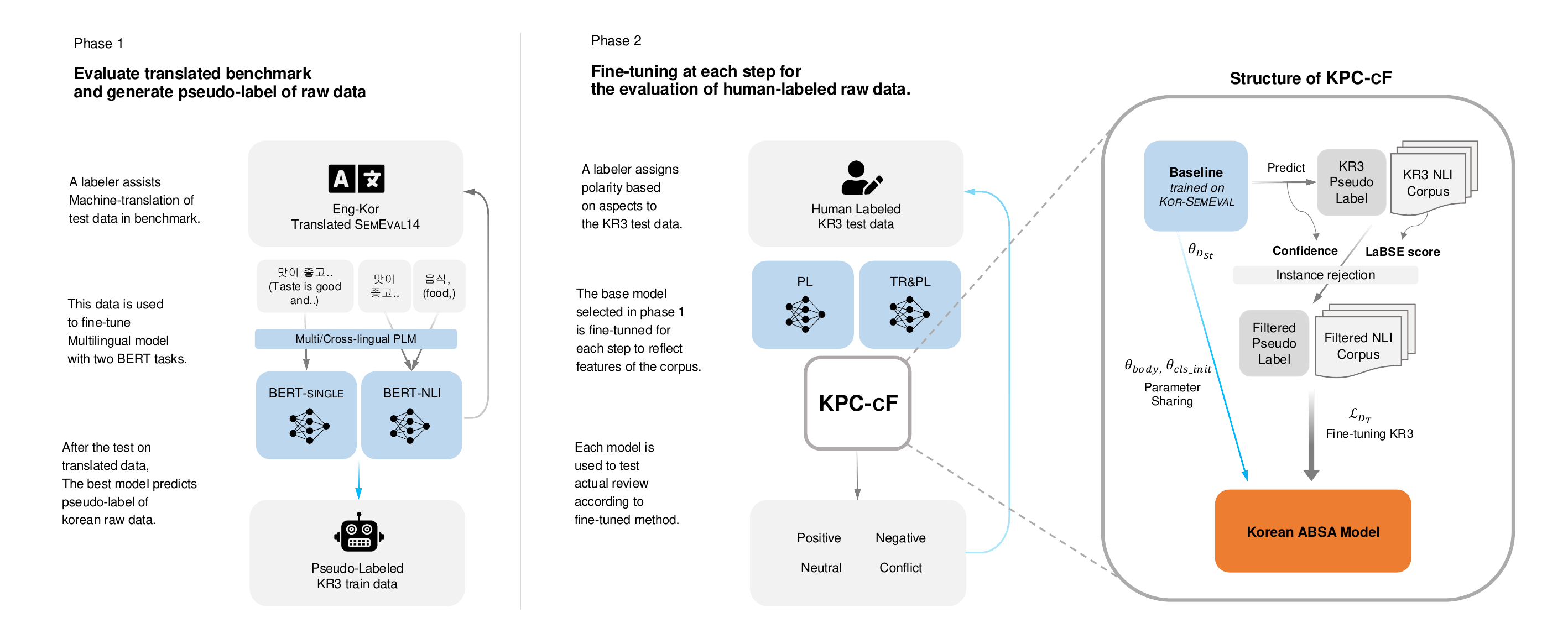}}
\caption{ A diagram illustrating the two phase of our method: (1) Fine-tuning Kor-SemEval and generate pseudo-labeled KR3, (2) Fine-tuning KR3 using baseline model selected phase 1. We illustrated the filtering process (right) for fine-tuning KR3 data. Blue arrows (left \& middle) indicate that this model is used to predict best label of review.}
\label{fig1}
\end{figure*}
In this study, we derive pseudo-labels for real Korean reviews using machine-translated English ABSA data, inspired by the past research \cite{balahur-turchi-2012-multilingual, hoshino2024cross}.
Moreover, we employ Dual filtering on the actual Korean corpus converted to implicit NLI task \cite{hendrycks2016baseline, sun-etal-2019-utilizing, feng-etal-2022-language}, thereby constructing an effective framework coined as \textbf{K}orean ABSA using \textbf{P}seudo-\textbf{\textsc{C}}lassifier with \textbf{C}orpus \textbf{F}iltering (\textbf{\textsc{KPC-cF}}) to achieve implicit-feature alignment. Through this process, we assess the impact of our constructed classifier on the practical alignment of actual reviews. 
We validate that the pseudo-classifier, generated through the sentence-pair approach, outperforms the single approach in translation task.
Furthermore, using the model that predicts the translated dataset most effectively as a baseline, we generate pseudo-labels for actual data and conduct real-world testing of Korean ABSA. This involves subsequent fine-tuning the filtered corpus based on language-agnostic embedding similarity for review and aspect sentence pairs, along with setting a threshold for Maximum Softmax Probability (\MSP) in pseudo-labels.

The main contributions of our work are:
\vspace{-5pt}
%{Large bullets: \Large \textbullet}
\begin{itemize}%[label={\large\textbullet}]
\item This is, to our knowledge, the first approach to generating a pseudo-classifier for automatic classification of aspect-based sentiment in the actual Korean domain.
%\item We show insights into the selecting and fine-tuning PLMs for effective Korean ABSA.
\vspace{-5pt}
\item For actual review-based ABSA, we propose a filtered NLI corpus as implicit feature and fine-tuning framework that enables important data selection in low-resource languages on models trained with high-resource dataset.
\vspace{-5pt}
\item A new challenging dataset of Korean ABSA, along with a KR3 and translated benchmark correlated with cross-lingual understanding.
\end{itemize}
%\vspace{-5pt}
\section{Task Description}\label{Section B.1}
In ABSA, \citet{sun-etal-2019-utilizing} set the task as equivalent to learning subtasks 3 (Aspect Category Detection) and 4 (Aspect Category Polarity) of SemEval-2014 Task 4 simultaneously. A similar approach was adopted for Korean ABSA in industrial reviews, aiming to develop a task-specific model through comparison of two PLMs (see Appendix\S\ref{Section B}), differing only in tokenization, vocabulary size, and model parameters \cite{conneau-etal-2020-unsupervised}. Defining a unified-serving model using multi-label-multi-class classification from a task-oriented perspective was considered impractical due to challenges in modifying pre-training set and the ongoing injection of additional data, rendering implicit mapping unattainable \cite{sun2020lamal, ahmed2022bert, qin2022lfpt, kossen2024context}. Consequently, problem has been redefined into BERT-based tasks (see Appendix\S\ref{Section B.1}).

\section{Two phase of Pseudo-Classifier}
\subsection{Motivation and Contribution}\label{Section 3.1}
The goal of this research is to propose a framework for achieving the best ABSA on actual data with Korean nuances through high-resource languages. Past research by \citet{balahur-turchi-2012-multilingual} has shown that Machine Translation (MT) systems can obtain training data for languages other than English in general sentiment classification. However, \citet{li2020unsupervised, zhang-etal-2021-cross} use explicitly aligned translation features, limiting the exploration of discriminative real target features in low-resource languages like Korean \cite{fei2022robustness, zhao2024systematic}.
%However, existing research using translation data for alignment and alignment-free methods \cite{li2020unsupervised, zhang-etal-2021-cross} inadequately address the challenge of universal knowledge transfer for linguistic subgroups like Korean. 
Also, although it was a different domain at 
\citet{zhou2021adaptive}, we found it necessary to investigate whether the concept of pseudo-labels could help \textbf{bridge the gap of feature $\Phi$ between translated source data $\bm{{D_S}_t}$ and actual target language data $\bm{D_T}$.}
Therefore, we attempted the following two phases to assess the impact of the generated pseudo-classifier, fine-tuned using translated datasets from the ABSA benchmark and pseudo-labeled actual review data, on Korean ABSA.
Figure \ref{fig1} shows the two-phase pseudo-classifiers we will employ. 
In the first phase, we investigate the data structure exhibiting performance aligned with ${D_S}_t$ by thoroughly learning the fine patterns of the labeled benchmark for each model. NLI structure is referenced based on our research domain and language settings, following the prior work \cite{hoshino-etal-2024-cross-lingual}.
In Phase 2, we fine-tune the baseline $\Psi_{pre}(\Phi({D_S}_t);\theta_{{D_S}_t})$, which was effective in training on ${D_S}_t$, by additionally incorporating pseudo-labeled $D_T$. Employing the tuned model $\Psi_{post}(\Phi(D_T);\theta_{{{D_S}_t}\rightarrow D_T})$, we conduct predictions and evaluations on manually labeled actual Korean reviews. Throughout this process, Dual filtering is performed to enhance implicit features $\Phi(D_T)$. %Detail of Language Adaptation in ABSA is provided in Appendix\S\ref{Section A.1.1}.
Detailed description of our training strategy is provided in Section\S\ref{Section 3.4}, Appendix\S\ref{Section C.0}.

\begin{algorithm}[t]
\caption{Dual Filtering}
\label{alg:dual_filtering}
\begin{algorithmic}[1]
\STATE \textbf{function} \textsc{Sampling}($Target$, $\Psi_{pre}$)
%\FUCTION{Sampling}{$Target$, $LM_{TR}$}
\STATE \quad \textbf{for} $i=1$ to Target \textbf{do}
%\FOR{$i=1$ {\bfseries to} Target}
    \STATE \quad\quad $x_s, x_a \gets$ Target sample $i$\qquad\quad\;\;\; {\scriptsize $\triangleright\, Attach\,aspect$}
    \STATE \quad\quad $score_{L} \gets$ LaBSE($x_s, x_a$)
    \STATE \quad\quad Add $score_{L}$ to $temp$
    \STATE \quad\quad $y \gets$ $\Psi_{pre}$($x_s, x_a$)\qquad\quad\quad\;\, {\scriptsize $\triangleright\, Model\,from\,phase\,1$}
    \STATE \quad\quad $score_{MSP} \gets$ $\MSP_{\Psi_{pre}}$($y\mid x_s, x_a$)
    \STATE \quad\quad \textbf{if} $score_{MSP} > threshold_{\tau_1}$
        \STATE \quad\quad\quad Add $(x_s, x_a, y, score_{L})$ to $batch$
    \STATE \quad\quad \textbf{end if}
\STATE \quad \textbf{end for}
\STATE \quad $avg_{\tau_2} \gets\frac{1}{N} \sum_{i=1}^{N} score_{L}[i]$ in $temp$
\STATE \quad $batch \gets batch$ [$batch.score_{L} > avg_{\tau_2}$]
\STATE \quad \textbf{return} $\Phi(batch)$ for $\Psi_{post,\, joint}$ Tuning
\STATE \textbf{end function}
%\ENDFUNCTIONLaBSE based
\end{algorithmic}
\end{algorithm}
\vspace{-10pt}  
%\subsection{Target-aligned Filtering}\label{AA}
\subsection{LaBSE based Filtering}\label{AA}
Our main approach, we aim to extract good-quality sentences-pair from the pseudo-NLI corpus. Language Agnostic BERT Sentence Embedding model \cite{feng-etal-2022-language} is a multilingual embedding model that supports 109 languages, including some Korean languages. 
\citet{feng-etal-2022-language} suggested that the dual-encoder architecture of the LaBSE model, originally designed for machine translation in source-target language data \cite{batheja-bhattacharyya-2022-improving, batheja-bhattacharyya-2023-little}, can be applied not only to other monolingual tasks like Semantic Textual Similarity (STS) but also to data (i.e., sentence pair-set) filtering for creating high-quality training corpora in terms of meaning equivalence. 
Therefore, to mitigate performance degradation caused by the linguistic gap between translated ${{D_S}_t}$ and actual Korean ${D_T}$ during fine-tuning, we introduce the following filtering method that enables the identification of meaning equivalence or connotation \cite{ghadery2019licd} in actual Korean sentence-pairs, even when viewed from the perspective of model trained on bilingual translation pairs. We generate the sentence embeddings for the review text and aspect of the pseudo-NLI corpora using the LaBSE model. Then, we compute the cosine similarity between the review text and aspect sentence embeddings. After that, we extract good quality NLI sentences based on a threshold value of the similarity scores. %We calculate the average similarity score on a  dataset from the our Target (KR3) NLI corpus. Our processed corpus consists of high-quality sentence pairs, so it helps us decide upon the threshold value.
\vspace{-10pt}  
\paragraph{LaBSE scoring} \quad Let $D_T = \{(x_s^i, x_a^i)\}_{i=1}^{N}$ be a pseudo-NLI corpus with $N$ examples, where ${x_s^i}$ and ${x_a^i}$ represents $i^{th}$ review and aspect sentence respectively. We first feed all the review  sentences present in the pseudo-parallel corpus as input to the LaBSE model\footnote{\url{https://huggingface.co/sentence-transformers/LaBSE}}, which is a Dual encoder model with BERT-based encoding modules to obtain review sentence embeddings ($S_i$). The sentence embeddings are extracted as the l2 normalized \texttt{[CLS]} token representations from the last
transformer block. Then, we feed all the aspect sentences as input to the LaBSE model to obtain aspect sentence embeddings ($A_i$). We then compute cosine similarity $(score_i)$ between the review and the corresponding aspect sentence embeddings.
\begin{equation}
S_i=LaBSE\left(x_s^i\right) 
\end{equation}
\begin{equation}
A_i=LaBSE\left(x_a^i\right) 
\end{equation}
\begin{equation}
score_i=cosine\_similarity\left(S_i, A_i\right)    
\end{equation}
We aimed to apply the LaBSE scoring to the actual Korean dataset $D_T$, KR3, intending to facilitate flexible learning compared to the Eng-Kor translated dataset ${D_S}_t$, Kor-SemEval (see Section\S\ref{Section C}).
\vspace{-10pt}    
\subsection{Confidence score Filtering}
Meanwhile, we need to develop a classifier $\Psi_{post}$ capable of optimal predictions on the $D_T$ test, which can be considered as out-of-distribution data separate from the ${D_S}_t$.
Drawing on previous research \cite{arora-etal-2021-types}, we expect that language shifts (i.e., ${D_S}_t \leftrightarrow D_T$) embody both Background and Semantic shift characteristics. %embody similar characteristics to Background and Semantic shift.
To ensure robust learning in both aspect detection and sentiment classification, we introduce additional thresholding on Maximum Softmax Probability (\MSP; \citeauthor{hendrycks2016baseline} \citeyear{hendrycks2016baseline}) after LaBSE-based filtering on the $D_T$ train set. When considering an input $x = (x_s, x_a) \in\sX$ and its corresponding pseudo-label $y\in\sY$, the score $s(x)$ for \MSP is expressed as:
%\begin{equation}
\begin{align}
\label{eqn:msp}
    s_{\MSP}(x) = \max_{k\in\sY} p_{\Psi_{\text{pre}}}(\hat{y}=k\mid x).
\end{align}
%\end{equation}
Through this, we intended a dual scoring and filtering process to ensure that our $\Psi_{post}$ does not retrain on misplaced confidence or subpar prediction outcomes for out-of-distribution data. 
This process is defined by the following expected success rate of the trained models, where $\mathcal{F}_{\text{Dual}}$ filters the sentence pairs and pseudo-labels. 
\begin{equation}
P_{\Psi_{\text{post}}} = \mathbb{E}_{(x_s, x_a, \hat{y}) \sim D_T} \left[ \mathcal{F}_{\text{Dual}}(x_s, x_a, \hat{y}) \right]
\end{equation}
Our algorithm for calculate scores and filter with the target $D_T$ batch set can be found in Algorithm \ref{alg:dual_filtering}.

\subsection{Target-aligned Objective}\label{Section 3.4}
In this section, we focus on the langauge adaptation of ABSA. The input data includes a set of labeled sentences from a translated source data ${D_S}_t = \{(x_s^h, x_a^h, y^h)\}_{h=1}^{N}$ and a set of unlabeled sentences from a target language data $D_T = \{(x_s^i, x_a^i)\}_{i=1}^{N}$. Our goal is to validate our filtering technique across three types of tasks: (1) training only pseudo-labeled $D_T$, (2) joint training using both ${D_S}_t$ and $D_T$, (3) transfer learning from ${D_S}_t$ to $D_T$.
Namely, unlike \citet{chen2023improving}, we maintain token embeddings and fine-tuned the transformer body \& head. Contrary to \citet{gururangan-etal-2020-dont}, we trained with ${D_S}_t$ and then reinitialized the classifier layer for further fine-tuning. This approach examines if the universal encoder and params $\theta$ aligned with ${D_S}_t$ enhance implicit alignment via pseudo-labels of refined $D_T$. 
\[
\mathcal{L}_{pre} = - \frac{1}{N} \sum_{h=1}^{N} \mathbb{E}_{(x^{(h)}, y^{(h)}) \sim {D_S}_t} \left[ \log p(y^{(h)} | x^{(h)}; \theta) \right] %\tag{6}
\]
\vspace{-20pt}
\begin{align}
\mathcal{L}_{post} \! &= \! - \frac{1}{N} \sum_{i=1}^{N} \mathbb{E}_{p(x^{(i)}, \hat{y}^{(i)})}\left[ \log p(\hat{y}^{(i)} | x^{(i)}; \theta') \! \cdot \mathbb{I} (\hat{s}^{(i)} \geq \tau) \right] \nonumber %\tag{7}
\end{align}
\vspace{-20pt}
\begin{align}
\mathcal{L}_{align} = \mathcal{L}_{pre} + \mathcal{L}_{post} %\tag{8}
\end{align}
\vspace{-15pt}
\begin{align}
\theta_{D_T} \leftarrow \theta' - \hat{s} \cdot \nabla_{\theta'} \mathcal{L}_{post} \quad \text{where}\;\; \theta' = \theta_{body} + \theta_{cls}^{\text{init}}
\end{align}
$\Psi_{\text{joint}}(\Phi({D_S}_t \sqcup D_T);\theta_{{D_S}_t \sqcup D_T})$ was also evaluated, despite its inefficiency in terms of memory and computation, for data validation purposes. Thus, shuffling was used in $\Psi_{\text{joint}}$.
\begin{assumption}[\textit{Contextual,} \textit{Sementical} gradient]
%\paragraph{Approximation of \theta_{\text{Context, Sementic}}} 
Reparameterization of \citet{goyal2024context} in Appendix\S\ref{Section A} , At timestep $t = i$, gradient of expected loss with respect to $W_{KQ,H}$ observes
    \begin{align}
        \theta_C^\top[- \nabla_{W_{KQ}}\mathcal{L}_{align}(W^{(i)})] \Phi(D_T \cdot \mathbb{I} (\hat{s}^{(i)} \geq \tau_{1,2})) > 0, \nonumber \\ 
        \theta_S^\top[- \nabla_{W_{H}}\mathcal{L}_{align}(W^{(i)})] \Phi(D_T \cdot \mathbb{I} (\hat{s}^{(i)} \geq \tau_{1,2})) > 0. %\nonumber\\ 
        %\theta^\top[- \nabla_{W_{H}}\mathcal{L}_{align}(W^{(h)})] \Phi({D_S}_t) < 0.
    \end{align}
\end{assumption}
\newcommand{\softmax}[1]{\sigma\left( #1 \right)}
\begin{assumption}[Target-Aligned Feature]
\label{prop:fact_mem}
Under Assumptions in objective, for any example $(x_s, x_a) \in$ $D_T$, after the gradient step at timestep $t = i$, the value embedding of the sentence-pair tokens is more predictive of the label $y$.
\begin{align}
    \softmax{W_H^{(i) \top} W_V^{(i)} \Phi(x)}_y -\softmax{W_H^{(h) \top} W_V^{(h)} \Phi(x)}_y  >  0
\end{align}
\end{assumption}
\vspace{-22pt}
\section{Experimental Setup}
%\subsection{Datasets}\label{AA}
\subsection{Dataset for Fine-Tuning and Test}\label{Section C}
\vspace{-5pt}
\paragraph{Kor-SemEval}
%SemEval 설명 보충해야함
We translate the SemEval-2014 Task 4 \cite{pontiki-etal-2014-semeval} dataset\footnote{\url{http://alt.qcri.org/semeval2014/task4/}}. Moreover, it is evaluated for Korean ABSA. 
The training data was machine-translated (by Google Translate), and Test data was corrected manually only for fewer than 10 instances where abnormal translations occurred after machine translation. Each sentence contains a list of aspect ${x_a}$
with the sentiment polarity ${y}$. Ultimately, given a
sentence ${x_s}$ in the sentence, we need to:
%\vspace{-5pt}
\begin{itemize}
\item detects the mention of an aspect ${x_a}$;
\vspace{-5pt}
\item determines the positive or negative sentiment polarity ${y}$ for the detected aspect.
\end{itemize}
\vspace{-5pt}
This setting allows us to jointly evaluate Subtask 3 (Aspect Category Detection) and 4 (Aspect Category Polarity).

  \begin{table*}[t!]
        \setlength{\tabcolsep}{9pt}
        \renewcommand{\arraystretch}{1}
	\centering\resizebox{\linewidth}{!}
        {
		\begin{tabular}{l l c c c c c c c c}
			\toprule
			\multirow{2}*{\textbf{Model}} & \;\;\,\textbf{\#Training} & \multirow{2}*{\textbf{Task Type}} & \multicolumn{3}{c}{\textbf{Aspect Category}} & & \multicolumn{3}{c}{\textbf{Polarity}}\\
			\cline{4-6}
			\cline{8-10}
			~ & \textbf{Storage/Tokens} & ~ & Precision & Recall & Micro-F1 & & 4-way acc & 3-way acc & Binary\\
                \midrule
                %\hline
                GPT-4o mini \cite{openai2024gpt4omini} & \qquad - \qquad - & - & 28.00 & 99.91\textsuperscript{*}& 43.74 & & 84.41 & 86.13 & 89.10 \\
                Babel-9B-Chat \cite{zhao2025babel} & \qquad - \qquad - & - & 64.31 & 87.88 & 74.27 & & 75.66 & 76.57 & 79.37 \\
                \textsc{Zero-Shot} \cite{li2020unsupervised} & 1.43MB 0.38M & $D_S$ & 75.31 & 74.41 & 74.15 & & 81.52 & 84.00 & 88.73 \\
                \textsc{Trans-Ta} \cite{zhang-etal-2021-cross} & 1.54MB 0.37M & ${D_S}_t$ &94.04\textsuperscript{*}& 78.94 & 85.78 & & 74.84 & 77.37 & 80.75 \\
   		    \midrule
                %\midrule
                Baseline\textsubscript{mBERT}\textsc{+PL}  & 4.74MB 1.28M & $D_T$ & 89.71& 77.60& 83.21& & 77.96& 80.15& 84.01\\
			Baseline\textsubscript{mBERT}\textsc{+PL-cF} &2.53MB 0.68M & $D_T$ & 87.45& 76.28& 81.46& & \underline{78.49}& \underline{80.69}& \underline{84.70}\\
			Baseline\textsubscript{XLM-R}\textsc{+PL}  & 4.60MB 1.13M & $D_T$ & 90.58& 79.26& 84.54& & 83.36& 85.71& 89.88\\
			Baseline\textsubscript{XLM-R}\textsc{+PL-cF} &2.15MB 0.52M & $D_T$ & 89.16& 79.24& 83.89& & \underline{83.89}& \underline{86.25}& \underline{90.40}\\
   		\midrule
   			Baseline\textsubscript{mBERT}\textsc{+TR+PL} & 6.28MB 1.73M & ${D_S}_t \sqcup D_T$ & 92.15& 78.28& 84.64& & 80.45& 82.71& 86.57\\   			\baseline{Baseline\textsubscript{mBERT}\textsc{+TR+PL-cF}} & \baseline{4.07MB 1.13M} & \baseline{${D_S}_t \sqcup D_T$} & \baseline{90.15}& \baseline{\textbf{81.20}}& \baseline{\textbf{85.42}}& \baseline{} & \baseline{\textbf{80.88}}& \baseline{\textbf{83.15}}& \baseline{\textbf{86.96}}\\
   			Baseline\textsubscript{mBERT}\textsc{+TR+PL} & 6.28MB 1.73M & ${D_S}_t \rightarrow D_T$ & 91.08& 77.48& 83.73& & 78.37& 80.58& 84.60\\
   			\baseline{Baseline\textsubscript{mBERT}\textsc{+TR+PL-cF}} & \baseline{4.07MB 1.13M} & \baseline{${D_S}_t \rightarrow D_T$} & \baseline{90.03}& \baseline{\textbf{77.80}}& \baseline{83.47}& \baseline{} & \baseline{\textbf{78.94}}& \baseline{\textbf{81.16}}& \baseline{\textbf{85.09}}\\
                %\hline
                \midrule
   			Baseline\textsubscript{XLM-R}\textsc{+TR+PL} & 6.14MB 1.51M & ${D_S}_t \sqcup D_T$ & 91.88& 83.66& 87.57& & 83.43& 85.78& 89.34\\
			\baseline{\textbf{\textsc{KPC-cF} \dag}} & \baseline{3.69MB 0.89M} & \baseline{${D_S}_t \sqcup D_T$} & \baseline{\textbf{92.72}}& \baseline{83.30}& \baseline{\textbf{87.75}}& \baseline{} & \baseline{\textbf{83.52}}& \baseline{\textbf{85.90}}& \baseline{\textbf{89.56}}\\
   			Baseline\textsubscript{XLM-R}\textsc{+TR+PL} & 6.14MB 1.51M & ${D_S}_t \rightarrow D_T$ & 91.50& 84.23& 87.69& & 84.21& 86.57& 90.35\\
			\baseline{\textbf{\textsc{KPC-cF}}} & \baseline{3.69MB 0.89M} & \baseline{${D_S}_t \rightarrow D_T$} & \baseline{\textbf{92.46}}& \baseline{\textbf{84.53}}& \baseline{\textbf{88.29}}& \baseline{} & \baseline{\textbf{84.34}}& \baseline{\textbf{86.72}}& \baseline{\textbf{90.89}}\\
		\bottomrule
		\end{tabular}}
        %\vspace{-0.2cm}
		\caption{\label{tab5} KR3 test set results for Aspect Category Detection (middle) and Aspect Category Polarity (right). We reported the computational requirements as \textbf{\#Training}. Specifically, for Baseline+PL, it refers to the overall capacity of structured NLI data and total number of valid tokens from KR3 Train.
        The performances where filtered models outperform non-filtered models under the same training conditions are highlighted in \textbf{bold} \& \underline{underline}. 
        \textsc{KPC-cF}\dag, is jointly fine-tuned on the Kor-SemEval \& Filtered KR3 Train. 
        Task Type is represented by Section\S\ref{Section 3.4}.
        For GPT-4o mini, Babel-9B-Chat, to reflect diversity in few-shot, we report the average of two results measured with temperature 0 or 0.7 and top-p 1, extending \citet{wang2023chatgpt}. 
        \textsc{Zero-Shot}, \textsc{Trans-Ta}  \cite{li2020unsupervised, zhang-etal-2021-cross} were evaluated using our settings in Appendix\S\ref{Section C.0}.} 
  \end{table*}

\paragraph{KR3}
Unlike the domains previously used for Korean sentiment classification \cite{lee2020korean, yang2021transformer, ban2022survey}, Korean Restaurant Review with Ratings (KR3) is a comprehensive dataset encompassing various food service establishments, constructed from actual certified map reviews.
In the case of restaurant reviews, words and expressions that evaluate positive and negative are mainly included, and real users often infer what a restaurant is like by looking at its reviews. Accordingly, Jung et al.$^{3}$ constructed the KR3 dataset by crawling and preprocessing user reviews and star ratings of websites that collect broad food service information and ratings.
KR3 has 388,111 positive and 70,910 negative, providing a total of 459,021 data plus 182,741 unclassified data, and distributed to Hugging Face\footnote{\url{https://huggingface.co/datasets/leey4n/KR3}}.
We structured our training and test datasets to match the size of Kor-SemEval. Specifically, we addressed potential biases by randomly sampling indices from the original KR3, ensuring that evaluations for a specific restaurant were non-overlapping. Additionally, we maintained an even distribution of positive, negative, and neutral (ambiguous) classes, irrespective of the aspects indicated in the original KR3. This preprocessing step aimed to capture a comprehensive representation of sentiments across diverse attributes of sentences in the dataset. Subsequently, the data were configured to suit sentence pair classification. %(see Table \ref{tab1}). 
To allocate polarity labels for each aspect within the KR3 dataset, pseudo-labeling was conducted utilizing the optimal model identified during the Kor-SemEval performance evaluation. Pseudo-labels were assigned to the KR3 training data, and post pseudo-labeling, the test data underwent manual re-labeling by researchers. 
%\vspace{-1.5pt}

Table \ref{tab1} shows some Kor-SemEval and KR3 training data samples. In the case of KR3, the negative aspect is better reflected. Meanwhile, while Kor-SemEval gave neutrality to mediocre service, KR3 did not give neutrality to mediocre taste. While positive and negative data have been sufficiently accumulated and reflected, the tendency for a lack of neutral data can be confirmed in advance through some samples. %Table \ref{tab2} shows the statistics of the test sets for each dataset. 
We have organized both Kor-SemEval and KR3 data as open-source to facilitate their use in various training and evaluation scenarios.
\vspace{-2.5cm}
\section{Experiment}
\vspace{-0.05cm}

\subsection{Main Experiment: KR3 Test Set}
\vspace{-0.15cm}
On the other hand, based on the results from Kor-SemEval, we examined the dissimilarity specific to the trans-align task between mBERT and XLM-R\textsubscript{Base} (see Appendix\S\ref{Section D}). Accordingly, We opted for the Baseline-NLI approach (i.e., Baseline\textsubscript{mBERT, XLM-R}), which demonstrated the best performance, as the base model for Phase 2 (see Figure \ref{fig1}, Section\S\ref{Section 3.1}). 
We conducted evaluations on KR3 test data. 
\vspace{-0.35cm}
\subsection{Main Results}\label{Section 4.4}
\vspace{-0.25cm}

To investigate the effect of features $\Phi(D_T)$ for each corpus, we conduct baseline tuning comparisons between the PL and the PL-\textsc{cF} (see Table \ref{tab5}, \ref{tab4} for details). The variants of our tuning framework includes:
%\vspace{-0.25cm}

(1) \textbf{Baseline\textsubscript{PLM}+PL (Pseudo-Labeled data):} Fine-tuning the Baseline PLM in an NLI manner with pseudo-KR3.

(2) \textbf{Baseline\textsubscript{PLM}+PL-\textsc{cF} (Corpus Filtering):} Fine-tuning Baseline PLM with the NLI data obtained by rejecting instance from pseudo-KR3, where the threshold of \MSP{ \cite{hendrycks2016baseline}} is less than 0.5 and the cosine similarity between LaBSE embeddings is less than 0.15.

(3) \textbf{Baseline\textsubscript{PLM}+TR (TRanslated data)+PL:} Fine-tuning Baseline PLM in an NLI manner (pre-tuned or jointly fine-tuned with Kor-SemEval) using pseudo-KR3.

(4) \textbf{\textsc{KPC-cF} (Baseline\textsubscript{PLM}+TR+PL-\textsc{cF}):} Fine-tuning Baseline PLM in an NLI manner (pre-tuned or jointly fine-tuned with Kor-SemEval) using PL-\textsc{cF}.

Results on the KR3 test set are presented in Table \ref{tab5} and Figure \ref{fig2}. We find that the \textsc{KPC-cF} approach achieved adequately trained results in both subtasks for the actual korean data. The model pre-tuned with Kor-SemEval achieves the best performance in Aspect Category Detection (ACD).
For Aspect Category Polarity (ACP), it performs exceptionally well in the tuning of Pseudo-Labels, especially in the Binary setting. Filtered Pseudo-Labels preserve this characteristic well and amplify the performance of all metrics within ACP.

   	\begin{table}[t]
		\centering\resizebox{\linewidth}{!}{
		\begin{tabular}{l r c c c}
			\toprule
			\multirow{2}*{\textbf{Dataset}}& \multirow{2}*{\textbf{Count}} & \textbf{LaBSE. S}&\textbf{Pred. Prob}& \textbf{Corr.C}\\
                &  & (mean) & (mean) & (target none) \\
                %\hline
                \midrule
                %Kor-SemEval (\textbf{TR}) & 15.23K & - & -  \\
                %\multicolumn{4}{l}{\baseline{\emph{Total sentiment for each aspect in \textbf{Testset}}}}\\
                \textsc{PL} (Baseline\textsubscript{mBERT})& 15.23K & 0.22$\scriptstyle{\pm{0.08}}$ & 0.986$\scriptstyle{\pm{0.05}}$ & -0.08$\scriptstyle{ p < 0.05}$\\
                %\cline{2-3}
 			%\midrule 
                %\textsc{PL} $>$ \textsc{L\textsubscript{0.15}} $\cap$ \textsc{PL} $>$ \textsc{Th\textsubscript{0.5}} 
                \textsc{PL}$\scriptstyle{>\textsc{L\textsubscript{0.15}}}$ $\cap$ \textsc{PL}$\scriptstyle{> \textsc{Th\textsubscript{0.5}}}$  (Baseline\textsubscript{mBERT})& 6.77K & 0.24$\scriptstyle{\pm{0.06}}$& 0.988$\scriptstyle{\pm{0.04}}$& -0.05$\scriptstyle{ p < 0.05}$\\ 
                %\hline
                \midrule
                \textsc{PL} (Baseline\textsubscript{XLM-R})& 15.23K & 0.21$\scriptstyle{\pm{0.08}}$& 0.69$\scriptstyle{\pm{0.17}}$& -0.09$\scriptstyle{ p < 0.05}$\\
                \textsc{PL}$\scriptstyle{>\textsc{L\textsubscript{0.15}}}$ $\cap$ \textsc{PL}$\scriptstyle{> \textsc{Th\textsubscript{0.5}}}$  
                (Baseline\textsubscript{XLM-R}) & 6.08K & 0.24$\scriptstyle{\pm{0.06}}$& 0.74$\scriptstyle{\pm{0.14}}$& -0.12$\scriptstyle{ p < 0.05}$\\ %($\uparrow$)       
			\bottomrule
		\end{tabular}}
		\caption{\label{tab4}Number of instances, average scores, and standard deviations for 4 classes, as well as the correlation coefficient between two scores in the none class, for KR3 fine-tuning data. We set the model-specific dataset, \textsc{PL} and \textsc{PL-cF}.}
	\end{table}

\begin{figure*}[t]
\centerline{\includegraphics[width=\textwidth, height=10.5cm]{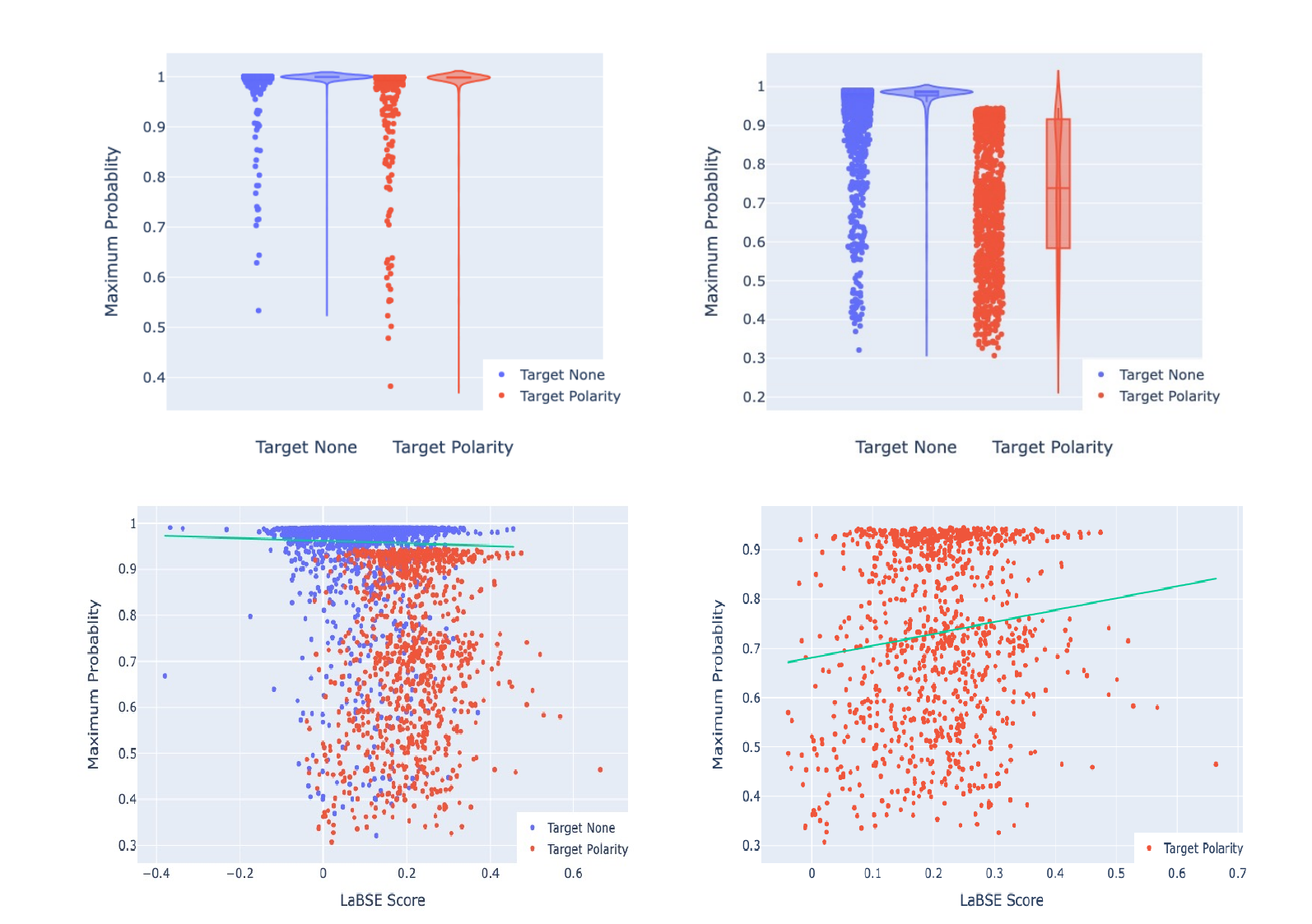}}
%\footnotesize\quad\quad \textbf{(a) TR w/o PL-\textsc{cF} \qquad\qquad\quad (b) TR w/ PL-\textsc{cF}}
\caption{\textbf{Top} - Maximum Probability Distribution of the Fine-Tuned Model, \textsc{KPC-cF} (left) vs. Baseline\textsubscript{XLM-R}+TR (right), \textbf{Bottom} - Maximum Probability Distribution of the Baseline\textsubscript{XLM-R}+TR with LaBSE Score Distribution, All classes (left) vs. 4 classes (right).}
\label{fig4}
\end{figure*}
\vspace{-0.25cm}
\section{Discussion}\label{Section A}
In Phase 1, XLM-R, known for its proficiency in capturing cross-lingual representations, exhibits an underfitting tendency concerning the contextual disparities in aspect vocabulary within a single task. This can be attributed to data scarcity relative to model availability for each classifier or viewed as a limitation in single task using SPM in low-resource Korean ABSA \cite{son2023removing}. Nevertheless, in the NLI task, it showcases potential by outperforming mBERT, guided by "aspect". Conversely, mBERT demonstrates stable results in both single and NLI tasks, exhibiting an overall accuracy increase, particularly in the NLI task (see Appendix\S\ref{Section D}). 
Furthermore, the pseudo-labels (PL) generated in Phase 2 refine polarity estimation, thereby enhancing classifier performance. Notably, PL-\textsc{cF} plays a crucial role in maintaining and improving accuracy and F1 score, even with fewer samples, unlike a simple addition of PL. The impact of further training on PL (PL further trained on TR) is evident through embedding analysis and ablation studies based on thresholding (see Figure \ref{fig3}, \ref{fig2}).  
Additionally, GPT-4o mini exhibits high recall but significantly lower precision, as it assigns polarity-labeled aspects to a much larger set. In contrast, \textsc{Trans-Ta} achieves higher precision, suggesting that alignment models trained on translated data can refine contextual categories relevant to domain classification. This further implies the importance of conservative evaluation in review classification, particularly for aspects that may impact reputation, as well as its significance in real-time inference.
  	\begin{table}[t]
		%\centering
		\centering\resizebox{1\linewidth}{!}{
        \fontsize{8.5}{11}\selectfont
		\begin{tabular}{l c c c}
			\toprule
			\multirow{2}*{\textbf{Model}} &\multirow{2}*{\textbf{Task Type}}& \textbf{KorSTS} & \textbf{KLUE-STS}\\
			%\cline{3-4}
                & & \multicolumn{2}{c}{Zero-shot Inference} \\
                \midrule
                %\multicolumn{3}{c}{\emph{Zero-shot evaluation}} \\
                Baseline\textsubscript{KLUE-R} & pre-trained & 23.78 & -10.91\\
                Baseline\textsubscript{KLUE-R-Large} & pre-trained & 13.33& -18.17\\
                \textsc{SetFit}\textsubscript{KLUE-R-13domain} & pre \& fine-tuned & 29.49& 8.37\\
                \midrule
                Baseline\textsubscript{XLM-R} & $\neg\,\textsc{sft}\,\,$ & 11.41& -1.23 \textsuperscript{*}\\
                Baseline\textsubscript{XLM-R}\textsc{+TR} &${D_S}_t$ & 10.26& 2.55 \textsuperscript{*}\\
                Baseline\textsubscript{XLM-R}\textsc{+\,PL} &$D_T$ & 12.18& 7.59\\
                %\cline{2-3}
			  Baseline\textsubscript{XLM-R}\textsc{+TR+PL} &${D_S}_t \rightarrow D_T$& 9.21& 6.35\\ 
                %\midrule 
                \baseline{\textbf{\textsc{KPC-cF}}} &\baseline{${D_S}_t \rightarrow D_T$}& 
 \baseline{\textbf{16.26}} & \baseline{\textbf{20.50}} \\
                %\cline{2-3}
 			%\midrule
 			%\hdashline 
			\bottomrule
		\end{tabular}}
		\caption{\label{tab11}\textbf{Contextual Implication} Comparisons of intermediate layer features as mean-pooled embeddings \cite{choi2021evaluation} on general STS. We report Spearman's \(\rho\) (\%). Higher is better. \textsuperscript{*}Values are for p $\approx$ 0.1; others within p $<$ 0.05. \textsc{SetFit} is selected based on the number of fine-tuning data.\tablefootnote{For details on \textsc{SetFit}\textsubscript{KLUE-R}, refer to \citet{tunstall2022efficient} and \url{https://huggingface.co/mini1013}.}}
	\end{table} 
    
Looking at Figure \ref{fig4}, compared to Baseline\textsubscript{XLM-R}+TR, \textsc{KPC-cF} significantly increases the maximum probability of Target Polarity through additional fine-tuning with filtered data. Despite a 40\% data reduction, it prevents biased class learning and maintains a balanced probability distribution across Target classes. Furthermore, for sentiment polarity data where the target is detected, higher LaBSE scores correspond to increased model confidence.  
\textsc{KPC-cF} also exhibits scalability in intermediate and shallow-layer feature comparisons. It enhances the model’s ability to capture general contextual relationships between sentence embeddings (see Table \ref{tab11}), as confirmed through zero-shot embedding evaluations using correlation coefficients against Ground Truth. Additionally, training a linear model with word-level features further validates the influence of filtered data characteristics on training outcomes (see Table \ref{delta0}). This confirms effectiveness of our approach in refining semantic label information. Our framework improves model stability and reliability by optimally rejecting uncertain or biased data.  

\begin{table}[t]
\centering{\resizebox{1\linewidth}{!}{
%\fontsize{8}{10}\selectfont
\begin{tabular}{lcccc}
\toprule
\multirow{2}*{\textbf{Model}} & \multicolumn{2}{c}{\textbf{Feature Span$^*$}} & \multicolumn{2}{c}{\textbf{TR+PL $\rightarrow$ \textbf{$\Delta$ TR+PL-\textsc{cF}}}}  \\
%\cline{2-3}
 & ${D_{S_t \sqcup T}}$ & ${x_{s \oplus a}}$ & Micro-F1 & 4-way acc  \\ 
\midrule
XLM-R\textsubscript{Base} I.T + LR & \checkmark & S.T & +16.01 & +6.29  \\
mBERT I.T + LR & \checkmark & S.T & +13.14 & +5.19 \\
TF-IDF I.T + LR & \checkmark & - & +0.19 & +0.19  \\
\bottomrule
\end{tabular}}}
\caption{\footnotesize \textbf{Sementic Stability} Performance variation of TR+PL vs. TR+PL-\textsc{cF} (generated by Baseline\textsubscript{XLM-R}) utilizing Logistic Regression as a shallow feature classifier for different Input Token IDs. $^*$We combined each Data \& NLI format w/w.o special tokens. To assess the contribution of previous ${D_S}_t$, shuffling was omitted.}
\label{delta0}
\end{table}

\subsection{Further Analysis}\label{Section 4.1}
To analyze the types of anticipatory effects that enable language models to exhibit improved performance in filtering corpora, we conducted a qualitative analysis by examining 116 cases where \textsc{KPC-cF} outputs matched Ground Truth but not those of the misaligned model (\textit{i.e.}, Baseline\textsubscript{XLM-R}\textsc{+TR+PL}). Interestingly, we identified the following four patterns that complement the prediction of misaligned model:
(1) \textbf{False Detection Refining}: misaligned model incorrectly predicted aspects not mentioned in the reviews with a specific polarity, but \textsc{KPC-cF} refined these to `None'.
(2) \textbf{Mismatch Refining}: misaligned model incorrectly predicted aspects mentioned in the reviews with a completely different polarity, which \textsc{KPC-cF} refined to the correct polarity.
(3) \textbf{Multi Context Detecting}: \textsc{KPC-cF} detected contexts within a review with multiple aspects, correctly assigning the `None' overlooked by misaligned model to the correct polarity.
(4) \textbf{Single Context Detecting}: \textsc{KPC-cF} detected contexts within a single-aspect review and correctly assigned the polarity that misaligned model had overlooked as `None'.

Patterns 1 and 2 emphasize the role of \textsc{KPC-cF} in providing error feedback. The filtered PL corpus indicates the presence of more significant polarities, guiding responses from the model. In contrast, patterns 3 and 4 demonstrate that the filtering framework allows the language model to utilize aligned aspect information. This adaptation of cohesive contexts from the translation corpus optimizes problem-solving for the target task. The frequency of each pattern from our qualitative assessment is provided in Table \ref{tab10}. On the other hand, given the somewhat limited effectiveness compared to Refining, the Detecting patterns suggest that training initiated from TR still possesses both scalability limits and potential. Therefore, we propose additional analysis and directions in Section\S\ref{Section 7}.

  	\begin{table}[t]
		\centering
		%\centering\resizebox{0.725\linewidth}{!}{
            \fontsize{8.5}{11}\selectfont
		\begin{tabular}{l c}
			\toprule
			\multirow{2}*{} &\textbf{Observed Instances} \%\\
                  & (\textsc{gt}-matched, not $\hat{y}_{\text{TR+PL}}$)\\
                \midrule
                False Detection Refining$^*$ &34.38\%\\
                %\cline{2-3}
 			%\midrule
			  Mismatch Refining &31.25\%\\  
                Multi \;Context Detecting &18.75\%\\
                %\cline{2-3}
 			%\midrule
			Single Context Detecting &15.62\%\\   
 			%\hdashline 
			\bottomrule
		\end{tabular}
		\caption{\label{tab10} Qualitative assessment results of \textsc{KPC-cF} predictions on the KR3 test set; $^*$We analyze aligned model outcomes with Refining and Detecting patterns.}
	\end{table}

\begin{figure*}[t]
\centerline{\includegraphics[width=\textwidth, height=5.25cm]{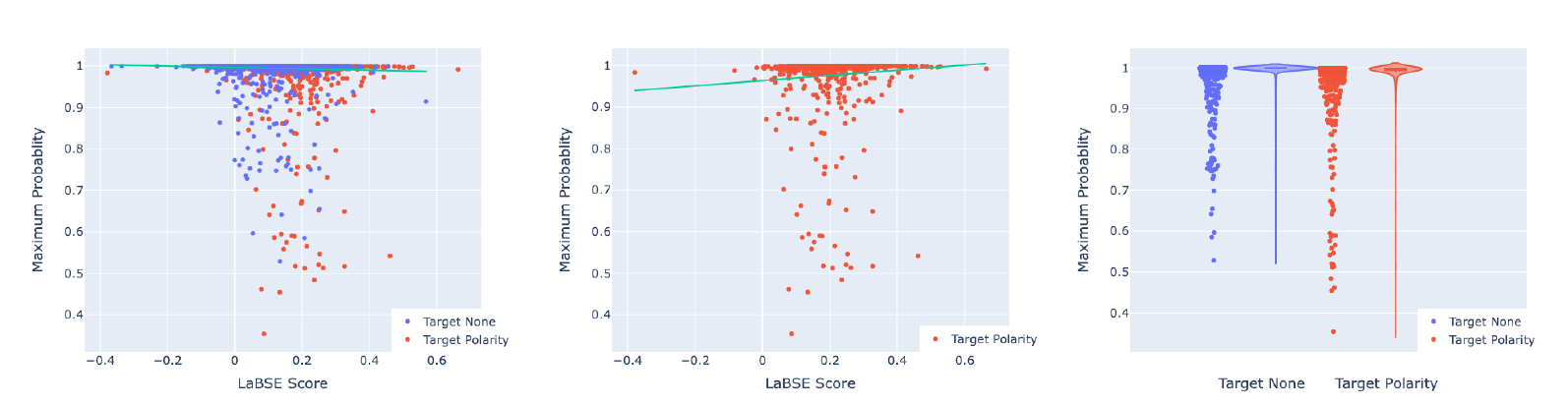}}
%\footnotesize\quad\quad \textbf{(a) TR w/o PL-\textsc{cF} \qquad\qquad\quad (b) TR w/ PL-\textsc{cF}}
\caption{\textbf{Left \& Middle} - Maximum Probability Distribution of the Baseline\textsubscript{mBERT}+TR with LaBSE Score Distribution, All classes (left) vs. 4 classes (middle), \textbf{Right} - Maximum Probability Distribution of the Fine-Tuned Model, Baseline\textsubscript{mBERT}+TR.}
\label{fig5}
\end{figure*}

\section{Limitations and Future Work}\label{Section 7}
Acknowledging limitations is crucial before delving into future work. While mBERT exhibits somewhat inadequate transfer effects, which can be attributed to its smaller model size. Due to pre-training solely on translation data, the scores are densely populated, resulting in limited performance improvement through additional training. 
Nevertheless, unlike jointly with Target None, the regression line of maximum probability for Target Polarity still trends upward according to LaBSE score (see Table \ref{tab4}, Figure \ref{fig5}), indicating that ongoing evaluation of threshold points for the two scores can be done amidst more diverse and precise patterns. Meanwhile, we conducted direct inference on individual NLI inputs through a demo\footnote{\url{https://huggingface.co/KorABSA}}. Upon detecting various aspects within the reviews, we observed a slight predominance towards the "None" class within the class-wise similar probability distributions  per sentence in TR training, while a more concentrated prediction tendency towards the binary class was observed in PL training. Furthermore, a tendency for one class to dominate over two or more aspect classes within review sentences was identified. This appears to stem from the class imbalance in TR and contextual differences \cite{jin2023kobbq} within all review data (i.e., TR \& PL).

Our analysis of the \textsc{PL-cF} thresholding reveals that while it reduces the number of predicted Target None classes resulting from extensive training on None class in \textsc{TR}, it also qualitatively explains why the error rate for Target None can be lower (see Table \ref{tab7}, \ref{tab8}). Contrary to experimental results that show it is reasonable to predict high-confidence Target Polarity labels when there is high surface similarity between sentences, cases where sentence similarity is less than 0.15 but predicted as Target None with confidence greater than 0.5 can hinder the model's understanding of deeper semantic relationships. For instance, issues related to hygiene and cleanliness are not detected through the overall atmosphere, and in cases where multiple aspects overlap, the model fails to distinguish them clearly, incorrectly labeling them all as None. Additionally, aspects such as waiting space, waiting time, and waiting staff, where the boundary between service and atmosphere is ambiguous or falls into general anecdotes, are consistently mislabeled as None.
Furthermore, our threshold filtering assigns a high-confidence Target None label for connotations with high similarity but lower errors, which supports a better distinction between semantic similarity and causal relationships. This is particularly effective when excluding personalized terms and Korean proper nouns, such as `Fixture (\textit{i.e.}, local native)' used by reviewers. Our current translation-based training does not sufficiently capture the tone of sentences composed of proper nouns and particles from various domains within Korea. %This limitation is evident not only in the qualitative analysis of specific data point errors but also in the KLUE-STS \cite{park2021klue} results in Tab. \ref{tab11}. 
However, we have confirmed through various evaluations (Table \ref{tab5}, \ref{tab4}, \ref{tab11}, \ref{delta0}, \ref{tab10}, \ref{tab8}, \ref{Ablation} and Figure \ref{fig2}, \ref{fig6}, \ref{fig7}, \ref{fig8}) that our filtering approach aids in learning actual level of entailment relationships, moving beyond merely identifying similar relationships between sentences.
To mitigate spurious correlations \cite{haig2003spurious, udomcharoenchaikit-etal-2022-mitigating}, we propose upper filtering that enables Target predictions to function as data facilitating the model’s learning of true entailment relationships between sentences, without relying solely on surface-level similarity in NLI pairs.

 	\begin{table}[t]
         . \begin{minipage}{.475\textwidth}
		\centering{\resizebox{\textwidth}{!}{
		\begin{tabular}{l c c}
			\toprule
			\multirow{2}*{\textbf{Dataset}} & \textbf{Count} & \textbf{Ratio} \%\\ %\textbf{Error}&\textbf{Ratio}\\
                  & (T.None) & (T.None/PL)\\
                %\hline
                \midrule                
                \textsc{PL}$\scriptstyle{>\textsc{L\textsubscript{0.15}}}$ $\cap$ \textsc{PL}$\scriptstyle{> \textsc{Th\textsubscript{0.5}}}$ (Baseline\textsubscript{mBERT})& 2891 &19.02\% $\downarrow$\\ %& 0.22 & 0.986\\
                %\cline{2-3}
 			%\midrule
			\textsc{PL}$\scriptstyle{<\textsc{L\textsubscript{0.15}}}$ $\cap$ \textsc{PL}$\scriptstyle{> \textsc{Th\textsubscript{0.5}}}$  (Baseline\textsubscript{mBERT})& 7634 & 50.12\%\\ %& 0.24 & 0.988 \\ 
                %\hline
                %\cdashline{1-3}
                \midrule
                \textsc{PL}$\scriptstyle{>\textsc{L\textsubscript{0.15}}}$ $\cap$ \textsc{PL}$\scriptstyle{> \textsc{Th\textsubscript{0.5}}}$ (Baseline\textsubscript{XLM-R}) & 2787 & 18.31\% $\downarrow$\\ %& 5 & 13.51\\
                %\cline{2-3}
 			%\midrule
			\textsc{PL}$\scriptstyle{<\textsc{L\textsubscript{0.15}}}$ $\cap$ \textsc{PL}$\scriptstyle{> \textsc{Th\textsubscript{0.5}}}$ (Baseline\textsubscript{XLM-R}) & 7470 & 49.06\%\\ %& 10 & 14.71 \\       
			\bottomrule
		\end{tabular}}}
            \vspace{-0.05cm}
		\caption{\label{tab7}Quality assessment results of PL filtering by two Baseline outputs; Target None ratio is 31.1\% and 30.75\% lower in the \textsc{PL-cF} threshold.}
	%\end{table}
          \end{minipage}%
         \hspace{0.0\textwidth} % Space between the two tables
         %\vspace{10pt}
          \begin{minipage}{.475\textwidth}
   	%\begin{table}[h]
		\centering{\resizebox{0.875\textwidth}{!}{
		\begin{tabular}{l c c}
			\toprule
			\multirow{2}*{\textbf{Dataset}}&\multicolumn{2}{c}{\textbf{Rejection Quality} \%}\\
                & High-Diff & General\\
                %&50&150\\
                %\hline
                \midrule                                \textsc{PL}$\scriptstyle{>\textsc{L\textsubscript{0.15}}}$ $\cap$ \textsc{PL}$\scriptstyle{> \textsc{Th\textsubscript{0.5}}}$ (Baseline\textsubscript{mBERT})& 16.00\% $\downarrow$& 17.95\% $\downarrow$\\
                %\cline{2-3}
 			%\midrule
			\textsc{PL}$\scriptstyle{<\textsc{L\textsubscript{0.15}}}$ $\cap$ \textsc{PL}$\scriptstyle{> \textsc{Th\textsubscript{0.5}}}$ (Baseline\textsubscript{mBERT})& 26.53\% & 21.65\%\\   
                \midrule               \textsc{PL}$\scriptstyle{>\textsc{L\textsubscript{0.15}}}$ $\cap$ \textsc{PL}$\scriptstyle{> \textsc{Th\textsubscript{0.5}}}$ (Baseline\textsubscript{XLM-R})& 14.00\% $\downarrow$& 16.45\% $\downarrow$\\
                %\cline{2-3}
 			%\midrule
			\textsc{PL}$\scriptstyle{<\textsc{L\textsubscript{0.15}}}$ $\cap$ \textsc{PL}$\scriptstyle{> \textsc{Th\textsubscript{0.5}}}$ (Baseline\textsubscript{XLM-R})& 25.00\% & 20.92\%\\ 

			\bottomrule
		\end{tabular}}}
            \vspace{-0.05cm}
		\caption{\label{tab8} Qualitative assessment results of PL filtering by two Baseline outputs; Target None errors decreased in both the 50 high-difference segments and the 150 general segments under \textsc{PL-cF} threshold.}
          \end{minipage}
	\end{table}

Thus, future research will delve into the association between embeddings from Bi and Cross-Encoder models, and explore alternative methods for filtering or joint representation of translated NLI set. Simultaneously, the KR3 test set effectively capture real-world imbalances, deliberately selecting sentences with a broader vocabulary is necessary for enhanced diversity. Fully constructing the KR3, it will serve as a benchmark for assessing LMs in Korean ABSA. Furthermore, Korean data from other domains should be acquired and evaluated. From an industrial perspective, our model could be integrated with multilingual AI and agents \cite{bi2024deepseek, openai2025operator} as a refining module. %Subsequently, detailed linguistic implications can be analyzed through cross-training and cross-evaluation with Kor-SemEval.
%\vspace{-0.5cm}
\section{Conclusion}
Aspect-Based Sentiment Analysis (ABSA) is a key subfield in text analytics, but the lack of high-quality labeled data hinders its industrial development. This study aims to address the language gap in ABSA.
By aligning implicit features through dual filtering, we propose an framework to align target languages solely through translation of existing benchmarks. This approach is expected to optimize tasks for both Korean and low-resource languages with limited pre-training on rich native data. Additionally, we presented Kor-SemEval, KR3 train (pseudo labeled \& filtered), and test data (Gold) composed of fine-grained set. We invite the community to extend Korean ABSA by providing new datasets, trained models, evaluation results, and metrics.

\section*{Acknowledgements}
We thank Chanseo Nam for manually correcting and annotating the Kor-SemEval and KR3 testset used to test our model, during his undergraduate research.
We also thank Kyeongpil Kang, Jinwuk Seok, Byungok Han, Woohan Yun, Guijin Son, Soyoung Yang and Joosun Yum for their valuable discussions. Thanks to DAVIAN, MODULABS and UST AI as well. This research was supported by Brian Impact Foundation, a non-profit organization dedicated to the advancement of science and technology for all.
\bibliography{example_paper}
\bibliographystyle{icml2024}

%%%%%%%%%%%%%%%%%%%%%%%%%%%%%%%%%%%%%%%%%%%%%%%%%%%%%%%%%%%%%%%%%%%%%%%%%%%%%%%
%%%%%%%%%%%%%%%%%%%%%%%%%%%%%%%%%%%%%%%%%%%%%%%%%%%%%%%%%%%%%%%%%%%%%%%%%%%%%%%
% APPENDIX
%%%%%%%%%%%%%%%%%%%%%%%%%%%%%%%%%%%%%%%%%%%%%%%%%%%%%%%%%%%%%%%%%%%%%%%%%%%%%%%
%%%%%%%%%%%%%%%%%%%%%%%%%%%%%%%%%%%%%%%%%%%%%%%%%%%%%%%%%%%%%%%%%%%%%%%%%%%%%%%
\newpage
\appendix
\onecolumn
\label{sec:appendix}
\section{Formulation under Assumption: Enhanced NLI-Based Pseudo-Classifier}\label{Section A}
%\section{Theoretical Analysis of Similarity, Confidence, and Parameter Updates}\label{Section A}

\subsection*{Similarity and Confidence}
The similarity between \(X_s\) (review tokens) and \(X_a\) (aspect tokens) is defined probabilistically:
\begin{equation}
\text{sim}(X_s, X_a) = \mathbb{E}_{p(X_s, X_a)} \left[ \frac{X_s \cdot X_a}{\|X_s\| \|X_a\|} \right],
\end{equation}
where \(p(X_s, X_a)\) represents the joint distribution of the input pairs. The confidence \(C\) of target polarity is derived as:
\begin{equation}
C = f\big(\text{sim}(X_s, X_a)\big),
\end{equation}
where \(f\) is a monotonically increasing function, such as \(f(x) = \log(1 + x)\) or \(f(x) = \sigma(ax + b)\).

\subsection*{Relationship with Sementic Label Accuracy}
The accuracy of pseudo label \(Y\), denoted as \(Acc_Y\), is modeled as an expectation \cite{arora-etal-2021-types} over the confidence:
\begin{equation}
Acc_Y = \mathbb{E}_{p(C)} [g(C)],
\end{equation}
where \(g(C)\) is a monotonically increasing function, e.g., \(g(C) = kC\) for \(k > 0\).

\subsection*{Loss Function and Gradients}
The model's loss \(\mathcal{L}(\theta)\) is expressed as:
\begin{equation}
\mathcal{L}(\theta) = \mathbb{E}_{p(X_s, X_a, Y)} \big[ \ell(Y, \hat{Y}) \cdot f(\text{sim}(X_s, X_a)) \big],
\end{equation}
where \(\ell(Y, \hat{Y})\) denotes the loss between the predicted label \(\hat{Y}\) and the pseudo label \(Y\). The gradients of the loss function are given by:
\begin{equation}
\nabla_\theta \mathcal{L}(\theta) = \mathbb{E}_{p(X_s, X_a, Y)} \big[ \nabla_\theta \ell(Y, \hat{Y}) \cdot f(\text{sim}(X_s, X_a)) \big].
\end{equation}

\subsection*{Updates to Model Parameters}
The output layer weights \(W_H\), responsible for sementic label-specific information, are updated as:
\begin{equation}
\Delta W_H = -\eta \nabla_{W_H} \mathcal{L}(W),
\end{equation}
where \(-\nabla_{W_H} \mathcal{L}(W)\) represents the gradient direction, and \(\eta\) is the learning rate. Similarly, the Key (\(W_{KQ}\)) weights are updated as:
\begin{equation}
\Delta W_{KQ} = -\eta \nabla_{W_{KQ}} \mathcal{L}(W).
\end{equation}

\subsection*{Gradient Propagation to Tokens}
The gradient propagation from the loss \(\mathcal{L}(\theta)\) affects the token embeddings \(X_s\) and \(X_a\), refining their representations during training:
\begin{equation}
\nabla_{X_s} = \frac{\partial \mathcal{L}(\theta)}{\partial X_s}, \quad \nabla_{X_a} = \frac{\partial \mathcal{L}(\theta)}{\partial X_a}.
\end{equation}
This ensures that the context similarity \(\text{sim}(X_s, X_a)\) evolves dynamically, aligning with the label \(Y\).

\subsection*{Alignment and Overall Dynamics}
The alignment of gradients and confidence in the context of label-specific updates can be expressed as:
\begin{equation}
\theta_{Context}^\top \big[ -\nabla_{W_{KQ}} \mathcal{L}_{align}(W) \big] \Phi(D_T \cdot \mathbb{I}(\hat{s} \geq \tau_{1,2})) > 0,
\end{equation}
\begin{equation}
\theta_{Sementic}^\top \big[ -\nabla_{W_H} \mathcal{L}_{align}(W) \big] \Phi(D_T \cdot \mathbb{I}(\hat{s} \geq \tau_{1,2})) > 0,
\end{equation}
where \(\mathbb{I}(\hat{s} \geq \tau_{1,2})\) is an indicator function, \(\Phi\) denotes a non-linear mapping, and \(D_T\) represents task-specific adjustments.

---

This formulation highlights the relationship between token similarity (\(X_s, X_a\)), parameter updates (\(W_H, W_{KQ}\)), and their alignment with confidence and task-specific refinements, providing insights into the dynamic learning process.

\section*{Extension of Experimental Results}\label{Section E}
\vspace{-12pt}
\begin{figure}[h]
\centerline{\includegraphics[width=1\textwidth]{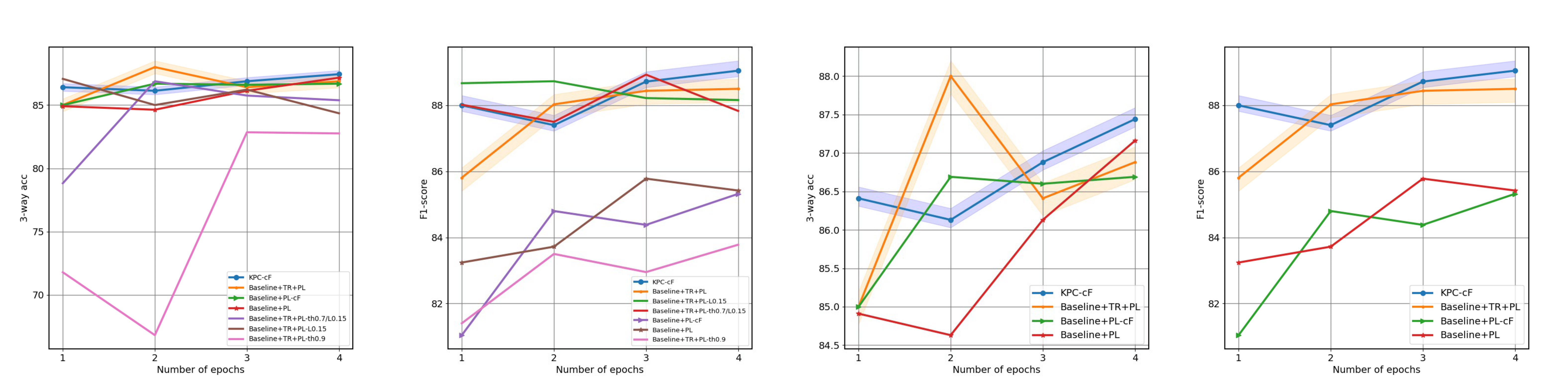}}
\caption{Performance of ACD and ACP on KR3 test set during adaptive fine-tuning ($D_T$ and ${D_S}_t \rightarrow D_T$). Left: results with the addition of other fine-tuned Baseline\textsubscript{XLM-R}. \textbf{th} denotes the threshold for confidence of pseudo-labeling, and \textbf{L} denotes the threshold for LaBSE filtering; Right: Baseline\textsubscript{XLM-R} tuning compared in this paper. Blue line represents \textsc{KPC-cF}.}
\label{fig2}
\end{figure}
\vspace{-10pt}

\section*{A.1 Gradient Stability and Directionality from Dual Scoring}

\textbf{Notation.} Let $x_s, x_a \in \mathbb{R}^d$ denote the sentence embeddings in a compact metric space $\mathcal{E}$. Let $\Phi(x_s, x_a) \in \mathcal{E} \subset \mathbb{R}^d$ be the fused input representation. Let $\theta = \{W_H, W_V, W_{KQ}\}$ be the trainable parameters. Define:
\begin{align*}
\text{sim}(x_s, x_a) &= \frac{\langle x_s, x_a \rangle}{\|x_s\| \cdot \|x_a\|} \,\in [-1,1], \\
C(x_s, x_a) &= \max_k \, p(k \mid x_s, x_a; \theta) \,\in [0,1], \\
\mathcal{D}_f &= \left\{(x_s, x_a, \hat{y}) \in \mathcal{D} \,\middle|\, \text{sim}(x_s, x_a) > \tau_1,\; C(x_s, x_a) > 0.5 \right\}
\end{align*}
where $\tau_1 = \mathbb{E}_{\mathcal{D}}[\text{sim}]$ is the empirical mean similarity. The fixed confidence threshold of $0.5$ reflects the theoretical minimum for non-random predictions and ensures robustness under calibration skewness.

Let the empirical loss be:
\[
\mathcal{L}(\theta) = \frac{1}{|\mathcal{D}_f|} \sum_{(x_s, x_a, \hat{y}) \in \mathcal{D}_f} \ell(f_\theta(x_s, x_a), \hat{y})
\]
where $\ell$ is the cross-entropy loss.

\textbf{Assumptions.} (1) $\mathcal{E}$ is compact. (2) $\text{sim}$ and $C$ are Lipschitz-continuous. (3) $\ell$ is locally convex and differentiable. 

\newcounter{manuallemma}
\refstepcounter{manuallemma}
\subsection*{Lemma \themanuallemma. Gradient Stability via Score Concentration}
Let $L$ be the Lipschitz constant of $\text{sim}$.
\[
|\text{sim}(x_1, x_2) - \text{sim}(x_1, x_3)| \leq L \cdot \|x_2 - x_3\|_2.
\]
After filtering, we observe in Table \ref{tab4}:
\begin{align*}
\mu_{\text{sim}}^{f} &:= \mathbb{E}_{\mathcal{D}_f}[\text{sim}] > \mathbb{E}_{\mathcal{D}}[\text{sim}] =: \mu_{\text{sim}}, \\
\sigma_{\text{sim}}^{f} &:= \sqrt{\mathbb{E}_{\mathcal{D}_f}[(\text{sim} - \mu_{\text{sim}}^{f})^2]} < \sqrt{\mathbb{E}_{\mathcal{D}}[(\text{sim} - \mu_{\text{sim}})^2]} =: \sigma_{\text{sim}}, \\
\mu_C^{f} &:= \mathbb{E}_{\mathcal{D}_f}[C] > \mathbb{E}_{\mathcal{D}}[C] =: \mu_C, \\
\sigma_C^{f} &:= \sqrt{\mathbb{E}_{\mathcal{D}_f}[(C - \mu_C^{f})^2]} < \sqrt{\mathbb{E}_{\mathcal{D}}[(C - \mu_C)^2]} =: \sigma_C.
\end{align*}
Define filtered gradient such that:
\[
\nabla_\theta \mathcal{L}_f = \frac{1}{|\mathcal{D}_f|} \sum_{(x_s, x_a, \hat{y}) \in \mathcal{D}_f} \nabla_\theta \ell(f_\theta(x_s, x_a), \hat{y})
\]
\vspace{-5pt}
Score concentration implies bounded gradient deviation:
\[
\forall (x_s, x_a) \in \mathcal{D}_f, \quad \mathbb{E}\left[\|\nabla_\theta \ell(f_\theta(x_s, x_a), \hat{y}) - \nabla_\theta \mathcal{L}_f\|_2^2\right] \leq \varepsilon_1(\sigma_{\text{sim}}^f, \sigma_C^f)
\]

\begin{proof}[Proof of Lemma 1]  % Gradient Stability via Score Concentration
Suppose, towards a contradiction, that the gradient variance on $\mathcal{D}_f$ is not bounded by the stated $\varepsilon_1$.  In particular, assume  
\[ 
\mathbb{E}_{(x_s,x_a)\sim \mathcal{D}_f}\Big[\big\|\nabla_\theta \ell(f_\theta(x_s,x_a),y) \;-\; \nabla_\theta L_f\big\|_2^2\Big] \;>\; \varepsilon_1(\sigma_f^{sim},\,\sigma_f^C)\,. 
\] 
Consider any two samples $i,j \in \mathcal{D}_f$. By Lipschitz continuity of $sim$ and $C$ and the differentiability of $\ell$, there exist constants $K_1,K_2$ such that the gradient difference is bounded by the score differences: 
\[ 
\big\|\nabla_\theta \ell(f_\theta(x_i),y_i) - \nabla_\theta \ell(f_\theta(x_j),y_j)\big\|^2 \;\le\; K_1\,|\,sim_i - sim_j\,|^2 \;+\; K_2\,|\,C_i - C_j\,|^2. 
\] 
Taking expectation over $i,j \sim \mathcal{D}_f$ gives an upper bound on the gradient variance: 
\[ 
\mathrm{Var}_{\mathcal{D}_f}[\nabla_\theta \ell] \;\le\; K_1\,(\sigma_f^{sim})^2 \;+\; K_2\,(\sigma_f^C)^2 \;=\; \varepsilon_1(\sigma_f^{sim},\,\sigma_f^C)\,. 
\] 
This contradicts the assumption that the variance exceeds $\varepsilon_1(\sigma_f^{sim},\sigma_f^C)$. Hence, the gradient deviation on $\mathcal{D}_f$ is bounded by $\varepsilon_1(\sigma_f^{sim}, \sigma_f^C)$ as claimed. 
\end{proof}

\refstepcounter{manuallemma}
\subsection*{Lemma \themanuallemma. Confidence-Guided Gradient Direction under Reinitialization}
Let the classifier weights be initialized as $W_H^{(t_0)} \sim \mathcal{N}(0, \sigma^2)$, and let $(x_s, x_a, y) \in \mathcal{D}_f$ be a filtered sample with high confidence $C(x_s, x_a) > \tau$. Then, at the training step $t = t_0$, where the classification head is reinitialized, the gradient satisfies:
\[
\left\langle \nabla_{W_H} \ell(f_{t_0}(x_s, x_a), y), \Phi(x_s, x_a) \right\rangle > 0,
\]
indicating that gradient updates are initially aligned with the semantic direction $\Phi(x_s, x_a)$. This alignment facilitates faster convergence during early fine-tuning. 

\begin{proof}[Proof of Lemma 2]  % Confidence-Guided Gradient Direction under Reinitialization
Assume for contradiction that the initial gradient is not aligned with the fused representation; i.e. $\langle\,\nabla_{W_H}\ell(f_{t0}(x_s,x_a),\,y),\; \Phi(x_s,x_a)\,\rangle \le 0$ at the reinitialization step $t_0$. Let $p = p(y \mid x_s,x_a; \theta(t_0))$ be the model’s predicted probability for class $y$ at $t_0$. Since $W_H(t_0)$ is random, $p<1$ for the true class $y$. The gradient of the loss w.r.t. the classification weight for class $y$ is 
\[ 
\nabla_{W_H,y}\,\ell(f_{t0}(x_s,x_a),y) \;=\; \big(p - 1\big)\,\Phi(x_s,x_a)\,, 
\] 
by the derivative of cross-entropy. The gradient descent update is $-\,\nabla_{W_H,y}\ell = (1-p)\,\Phi(x_s,x_a)$. Its inner product with $\Phi(x_s,x_a)$ is 
\[ 
\langle -\,\nabla_{W_H,y}\ell,\; \Phi(x_s,x_a)\rangle \;=\; (1-p)\,\|\Phi(x_s,x_a)\|^2 \;>\;0\,,
\] 
since $1-p>0$. This means the weight update at $t_0$ is positively aligned with $\Phi(x_s,x_a)$, i.e. $\langle \nabla_{W_H}\ell,\;\Phi\rangle > 0$.  This contradicts our assumption that $\langle \nabla_{W_H}\ell,\;\Phi\rangle \le 0$. Therefore, the initial gradient is indeed aligned with the semantic direction $\Phi(x_s,x_a)$, yielding $\langle \nabla_{W_H}\ell(f_{t0}(x_s,x_a),y),\; \Phi(x_s,x_a)\rangle > 0$. 
\end{proof}

Tables \ref{tab5} and \ref{delta0} show that, for semantic label detection, filtered data consistently yields higher ACP scores than unfiltered data. Even the Baseline\textsubscript{mBERT}+TR+PL-\textsc{cF} model—which exhibits a relatively low transfer learning effect in Table \ref{tab5}—achieves higher ACP with filtered data. Likewise, Baseline\textsc{+PL} variants without TR training also show improved ACP under filtering, further supporting our lemma.

\refstepcounter{manuallemma}
\label{lemma:compactness}
\subsection*{Lemma \themanuallemma. Compactness and Gradient Directionality}\label{lemma:compactness}

Let $\{\Phi(x_s^t, x_a^t)\}_{t=1}^\infty \subset \mathcal{E}$ be a sequence of fused representations, where $\mathcal{E}$ is the compact representation space defined in Assumption (1).

By the topological definition of compactness, for every open cover $\{U_\alpha\}_{\alpha \in A}$ of $\mathcal{E}$, there exists a finite subcover.  
Hence, the sequence $\{\Phi(x_s^t, x_a^t)\}$ cannot diverge or oscillate unboundedly, and must lie within a finite union of open sets.  
This guarantees the existence of a convergent subsequence:
\[
\Phi(x_s^{t_k}, x_a^{t_k}) \rightarrow \Phi^* \in \mathcal{E} \quad \text{as} \quad k \rightarrow \infty.
\]

Assuming $\ell$ is continuous in $\Phi$ and $f_\theta$ is differentiable, we obtain:
\[
\nabla_\theta \ell(f_\theta(\Phi(x_s^{t_k}, x_a^{t_k})), y_{t_k}) \rightarrow \nabla_\theta \ell(f_\theta(\Phi^*), y^*).
\]

Furthermore, let $q^{(t)} := W_Q \Phi(x_s^t, x_a^t)$ and $k^{(t)} := W_K \Phi(x_s^t, x_a^t)$ denote the query and key projections.  
By linearity and the convergence $\Phi(x_s^t, x_a^t) \to \Phi^*$, we have:
\[
\langle q^{(t)}, k^{(t)} \rangle = \langle W_Q \Phi(x_s^t, x_a^t), W_K \Phi(x_s^t, x_a^t) \rangle
\rightarrow \langle W_Q \Phi^*, W_K \Phi^* \rangle.
\]

Therefore, the attention score matrix defined by
\[
A^{(t)} := \text{softmax}\left(\frac{q^{(t)} (k^{(t)})^\top}{\sqrt{d_k}} \right)
\]
converges to
\[
A^* := \text{softmax}\left(\frac{W_Q \Phi^* (W_K \Phi^*)^\top}{\sqrt{d_k}} \right),
\]
and the attention structure becomes stable across samples.

Finally, the gradient w.r.t. $W_{KQ}$ also converges:
\[
\nabla_{W_{KQ}} \ell(f_\theta(\Phi(x_s^t, x_a^t)), y_t)
\rightarrow \nabla_{W_{KQ}} \ell(f_\theta(\Phi^*), y^*),
\]
with directional alignment preserved:
\[
\left\langle \nabla_{W_{KQ}} \mathcal{L}_f,\; W_{KQ} \Phi^* \right\rangle > 0,
\]
indicating alignment with Assumption 3.1, Figure \ref{fig6}, \ref{fig7}, \ref{fig8}, Table \ref{tab11} and enhanced ACD score in Table \ref{tab5}.

\qed

\subsection*{Lemma 4. Logit Monotonicity Post-Filtering}
Let $\sigma$ denote softmax and define logit scores:
\begin{align*}
P_{\text{pre}}(y|x_s, x_a) &= \sigma(W_H^{(t_0)\top} W_V^{(t_0)} \Phi(x_s, x_a))_y,\\ 
P_{\text{post}}(y|x_s, x_a) &= \sigma(W_H^{(t)\top} W_V^{(t)} \Phi(x_s, x_a))_y
\end{align*}
Then:
\[
P_{\text{post}}(y|x_s, x_a) - P_{\text{pre}}(y|x_s, x_a) > \delta > 0,
\]
indicating alignment with Assumption 3.2 and Figure \ref{fig4}.

\begin{proof}[Proof of Lemma 4]  % Logit Monotonicity Post-Filtering
Assume the fine-tuning yields no increase in the predicted probability for the true class $y$. Let $p_{\text{pre}} = P_{\text{pre}}(y \mid x_s,x_a)$ be the model’s probability for $y$ with the initial classifier $W_H(t_0)$, and $p_{\text{post}} = P_{\text{post}}(y \mid x_s,x_a)$ after training. Our assumption is $p_{\text{post}} \le p_{\text{pre}}$. However, from Lemma 2 we know the initial gradient update increases the logit for class $y$. In particular, after one gradient step of size $\eta$, the logit $z_y$ for class $y$ satisfies 
\[ 
z_y(t_0+\!1) \;=\; z_y(t_0)\;+\;\eta\,(1-p_{\text{pre}})\,\|\Phi(x_s,x_a)\|^2, 
\] 
so $z_y(t_0+1) > z_y(t_0)$ since $1-p_{\text{pre}}>0$. This implies $P(y \mid x_s,x_a)$ increases after the first update: $p^{(1)} := P(y|x_s,x_a)_{t_0+1} > p_{\text{pre}}$. By local convexity of $\ell$ (Assumption 3), further gradient steps will continue to improve the prediction on this sample. Therefore $p_{\text{post}} \ge p^{(1)} > p_{\text{pre}}$, contradicting the assumption $p_{\text{post}} \le p_{\text{pre}}$. We conclude $p_{\text{post}} - p_{\text{pre}} > 0$. In fact, one can take $\delta := p^{(1)} - p_{\text{pre}} > 0$ as a guaranteed improvement, so $P_{\text{post}}(y|x_s,x_a) - P_{\text{pre}}(y|x_s,x_a) > \delta$. 
\end{proof}

\subsection*{Theorem. Convergence under Low Gradient Dispersion}
Assume:
\[
\forall (x_s, x_a) \in \mathcal{D}_f, \quad \mathbb{E}\left[\|\nabla_\theta \mathcal{L} - \mathbb{E}[\nabla_\theta \mathcal{L}]\|^2\right] \leq \sigma^2
\]
Then for $L$-smooth $\mathcal{L}$ and AdamW optimizer:
\[
\lim_{t \to \infty} \left\| \frac{1}{|\mathcal{D}_f|} \sum_{(x_s, x_a, \hat{y})} \nabla_\theta \ell(f_\theta(x_s, x_a), \hat{y}) \right\|_2 \to 0
\]

\begin{proof}[Proof of Convergence Theorem]  % Convergence under Low Gradient Dispersion
Assume, to reach a contradiction, that $\|\nabla_\theta L(\theta_t)\|_2$ does not converge to $0$. Then there exists some $\epsilon > 0$ and an infinite sequence of iterations $\{t_k\}$ such that $\|\nabla_\theta L(\theta_{t_k})\|_2 \ge \epsilon$ for all $k$. Under the bounded gradient dispersion assumption, the Adam optimizer’s adaptive step size can be treated as a positive constant $\eta>0$ (for sufficiently large $t_k$). Because $L(\theta)$ is $L$-smooth, the descent lemma gives for a small enough $\eta$ (e.g. $\eta \le 1/L$): 
\[ 
L(\theta_{t+1}) \;\le\; L(\theta_t) \;-\; \eta\,\Big(1 - \frac{L\,\eta}{2}\Big)\,\|\nabla_\theta L(\theta_t)\|_2^2. 
\] 
Whenever $\|\nabla_\theta L(\theta_t)\|_2 \ge \epsilon$, this implies $L(\theta_{t+1}) \le L(\theta_t) - c\,\epsilon^2$ for some constant $c = \eta(1-\tfrac{L\eta}{2}) > 0$. In other words, every iteration with large gradient norm yields at least a fixed reduction $c\,\epsilon^2$ in the loss. After $N$ such iterations (with $\|\nabla L\| \ge \epsilon$), one would have 
\[ 
L(\theta_{t_N}) \;\le\; L(\theta_{t_0}) \;-\; N\,c\,\epsilon^2. 
\] 
For large $N$, the right-hand side becomes negative, which is impossible since $L(\theta)$ (being an average cross-entropy) is bounded below by $0$. This contradiction implies our assumption was false. Therefore $\|\nabla_\theta L(\theta_t)\|_2 \to 0$ as $t \to \infty$, i.e. the training converges to a stationary point. 
\end{proof}

\subsection*{Summary of Theoretical Guarantees}

The dual-filtered set $\mathcal{D}_f$ satisfies:
\begin{align*}
&\textbf{(Compactness)} \quad \Phi(x_s, x_a) \in \mathcal{E}_c \subset \mathcal{E},\quad \text{with } \mathcal{E}_c \text{ compact} \\
&\textbf{(Gradient stability)} \quad \|\nabla_\theta \ell(f_\theta(x_s, x_a), y) - \nabla_\theta \mathcal{L}_f\| \leq \varepsilon, \; \forall (x_s, x_a, y) \in \mathcal{D}_f \\
&\textbf{(Convergence)} \quad \left\| \nabla_\theta \mathcal{L}_f \right\|_2 \to 0 \quad \text{as } t \to \infty
\end{align*}

Under the NTK regime \cite{jacot2018neural}, these enhancements extend to shallow-layer generalization (Table \ref{delta0}), improving decision boundary sharpness.

\qed

\section*{\centering Preliminaries and Assumptions}\label{PA}

\noindent
Setting. Let \(\Phi(x_s, x_a) \in \mathbb{R}^d\) be a token-embedding feature for inputs \((x_s, x_a)\). 
We assume \(\Phi\) maps into some subset \(\mathcal{E} \subset \mathbb{R}^d\).

\subsection*{Assumption (1): \(\mathcal{E}\) is Compact}

Since \(\Phi(x_s,x_a) \in \mathcal{E}\) for all \((x_s,x_a)\), we want \(\mathcal{E}\) to be a compact subset of \(\mathbb{R}^d\). By the Heine--Borel theorem, compactness in \(\mathbb{R}^d\) is equivalent to being closed and bounded. Hence:

\begin{enumerate}
\item Boundedness. There exists \(R>0\) such that
\[
\|\Phi(x_s,x_a)\|_2 \;\le\; R
\quad
\text{for all }(x_s,x_a).
\]
\item Closedness. For any convergent sequence \(\{\Phi(x_s^n,x_a^n)\}_{n=1}^{\infty}\subset \mathcal{E}\), its limit also lies in \(\mathcal{E}\). Equivalently, \(\mathcal{E}\) contains all its limit points.
\end{enumerate}

\subsection*{Assumption (2): \(\mathrm{sim}\) and \(C\) are Lipschitz-Continuous}

Define
\[
\mathrm{sim}(x_s,x_a) \;=\; \frac{\langle x_s, x_a\rangle}{\|x_s\|\cdot\|x_a\|},
\quad
C(x_s,x_a) \;=\; \max_{k}\, p\bigl(k\mid x_s,x_a;\theta\bigr).
\]
Each must be Lipschitz in \(\Phi(x_s,x_a)\). Formally, there is an \(L_f>0\) such that
\[
\bigl|\,
f(x_s,x_a) - f(x_s',x_a')
\bigr|
\;\le\; 
L_f \,\|\Phi(x_s,x_a) - \Phi(x_s',x_a')\|_2,
\quad
\forall (x_s,x_a),(x_s',x_a')\in\mathcal{D}.
\]

\subsection*{Assumption (3): \(\ell\) is Locally Convex and Differentiable}

Let \(\ell\bigl(f_\theta(\Phi(x_s,x_a)),y\bigr)\) be a loss (e.g.\ cross-entropy). We require \(\ell\) to be differentiable, and in each local neighborhood it behaves like a convex function, ensuring standard gradient-based arguments near a minimum.

\section{Background and Related Work}\label{Section B}
\paragraph{Real-World ABSA}
Early research in ABSA primarily focused on single-element prediction, such as aspect term, category extraction \cite{ijcai16-ate, acl18-ate-xuhu, acl19-s2s-ate, naacl21-asap}, and sentiment polarity classification \cite{emnlp16-asc, emnlp18-asc, emnlp19-asc-category, emnlp20-asc}. After the advent of pre-trained language models, BERT-based architectures have become dominant in ABSA research~\cite{zhang2022survey,cai-2023-memdabsa,gao-etal-2022-lego-absa,wang-2024-unifiedABSA}. Additionally, text-to-text or LLM approaches \cite{gou-etal-2023-mvp, yang2024single} have been explored to improve ABSA tasks. These approaches leveraging pre-trained models assume high-performance extraction in settings with sufficient monolingual context and labeled data. However, in real-world multilingual  ABSA scenarios, challenges such as paraphrasing variations, semantic consistency with aspects, and polarity prediction quality \cite{zhang-etal-2021-cross, hsu-etal-2021-semantics, fei2022robustness} can arise, leading to performance degradation. 
To address these challenges in real-world scenarios, we employ an NLI-based pseudo-labeling and filtering approach. By inferring sentiment through structured and refined sentence relationships, our method mitigates semantic shifts and polarity distortions while adapting to the target data. This approach enhances (1) robust alignment between review-aspect context and sentiment information, even when word alignment is disrupted during translation, and (2) generalization performance in multilingual settings.
%\vspace{-5pt}
%\subsection{Additional information}\label{Section A.1}
\paragraph{mBERT}
Multilingual BERT is a BERT trained for multilingual tasks. It was trained on monolingual Wikipedia articles in 104 different languages. It is intended to enable mBERT finetuned in one language to make predictions for another. \citet{azhar2020fine} and \citet{jafarian2021exploiting} show that mbert performs effectively in a variety of multilingual Aspect-based sentiment analysis. It is also actively used as a base model in other tasks of Korean NLP \cite{lee2021korealbert, park2021klue}, but is rarely confirmed in  Korean ABSA tasks. Thus, our study used the pre-trained mBERT base model with 12 layers and 12 heads (i.e., 12 transfomer encoders). This model generates a 768-dimensional vector for each word. We used the 768-dimensional vector of the Extract layer to represent the comment. Like the English language subtasks, a single Dense layer was used as the classification model.
%\vspace{-5pt}
\paragraph{XLM-R}
XLM-RoBERTa \cite{conneau-etal-2020-unsupervised} is a cross-lingual model that aims to tackle the curse-of-multilingualism problem of cross-lingual models. It is inspired by RoBERTa \cite{liu2019roberta}, trained in up to 100 languages, and outperforms mBERT in multiple cross-lingual ABSA benchmarks \cite{zhang-etal-2021-cross, phan2021exploring, szolomicka2022multiaspectemo}. However, like mBERT, Korean ABSA has yet to be actively evaluated, so we used it as a base model. We use the base version (XLM-R\textsubscript{Base}) coupled with an attention head classifier, the same optimizer as mBERT. 

\subsection{Classification approach}\label{Section B.1}
%레퍼런스 달기 및 추가할거면 \cite{hyun-etal-2020-building}
%\vspace{-5pt}
\paragraph{Single sentence Classification}
BERT for single-sentence classification tasks. For ABSA, we fine-tune the pre-trained BERT model to train $n_a$ (i.e., number of aspect categories) classifiers for all aspects and then summarize the results. The input representation of the BERT can represent a pair of text sentences in a sequence of tokens. In the case of a single task, only one review text is tokenized and inputted as a sequence of tokens. A given token's input representation is constructed by summing the corresponding token, segment, and position embeddings. For classification tasks, the first word of each sequence is a unique classification embedding \texttt{[CLS]}. Segment embeddings in single sentence classification use one.
%\vspace{-5pt}
\paragraph{Sentence-pair Classification}
%레퍼런스 달기 및 추가할거면 \cite{sun-etal-2019-utilizing}
Based on the auxiliary sentence constructed as aspect word text, we use the sentence-pair classification approach to solve ABSA. The input representation is typically the same with the single-sentence approach. The difference is that we have to add two separator tokens \texttt{[SEP]}, the first placed between the last token of the first sentence and the first token of the second sentence. The other is placed at the end of the second sentence after its last token. This process uses both segment embeddings (In the case of XLM-RoBERTa, an additional one is placed in the first position, resulting in a total of three segment embeddings).
For the training phase in the sentence-pair classification approach, we only need to train one classifier to perform both aspect categorization and sentiment classification. Add one classification layer to the Transformer output and apply the softmax activation function.
%\vspace{-5pt}
\paragraph{Ensemble} Meanwhile, we additionally use a voting-based ensemble, a typical ensemble method. The ensemble can confirm generalized performance based on similarity of model results in NLI task \cite{xu2020improving}. So, we add separate power-mean ensemble result to identify a metric that amplifies probabilities based on the Pre-trained Language Models (PLMs). we reported the ensemble results of the top-performing models trained on NLI tasks for each PLM.

	\begin{table*}[t]
		\centering\resizebox{\textwidth}{!}{
            %\small
		\begin{tabular}{c l l}
			\toprule
			\textbf{Kor-SemEval Train}&\textbf{Aspect}& \textbf{Polarity}\\
                \midrule        
                \underline{서비스}는 평범했고 \underline{에어컨}이 없어서& 가격 (price)& 없음 (None)   \\
                %\cline{2-3}
                \underline{편안한 식사}를 할 수 없었습니다.&일화 (anecdotes)&없음 (None)    \\
                %\cline{2-3}
                ~ & 음식 (food)&없음 (None)   \\
                %\cline{2-3}
                (\emph{The \underline{service} was mediocre} \emph{and the lack of \underline{air conditioning}}& 분위기 (ambience)& \colorbox{pink}{부정} (\colorbox{pink}{Negative}) \\
                %\cline{2-3}
                \emph{made for a less than \underline{comfortable meal}.})& 서비스 (service)& \colorbox{light-blue}{중립} (\colorbox{light-blue}{Neutral}) \\
 			\midrule
			\textbf{KR3 Train}&\textbf{Aspect}& \textbf{Polarity}\\ 
 			\midrule
		      %\hline
                %\multicolumn{3}{r}{\baseline{\emph{Input form in NLI with \textbf{Pseudo-Label}}}}\\
		      \underline{가로수길}에서 조금 멀어요 \underline{점심시간에 대기} 엄청납니다 & 가격 (price)& 없음 (None)  \\
                %\cline{2-3}
		      일행 모두 있어야 들어갈 수 있어요 \underline{맛은 보통이에요}&일화 (anecdotes)& \colorbox{pink}{부정} (\colorbox{pink}{Negative}) \\
                %\cline{2-3}
		      ~ & 음식 (food)& \colorbox{light-blue}{없음} (\colorbox{light-blue}{None}) \\
                %\cline{2-3}
		      (\emph{It's a little far from \underline{Garosu-gil}. There's a huge \underline{wait during lunch time}.} & 분위기 (ambience)& \colorbox{pink}{부정} (\colorbox{pink}{Negative}) \\
                %\cline{2-3}
		      \emph{You have to have everyone in your group to get in.} \emph{The \underline{taste is average}.})& 서비스 (service)& 없음 (None) \\
			\bottomrule
		\end{tabular}}
		\caption{\label{tab1}Samples of Kor-SemEval and KR3 train dataset.} %For KR3 Train, the samples represent pseudo-labeled data.}
	\end{table*}
%\vspace{-5pt}
\section{Experimental Setup}\label{Section C.0}
\subsection{Hyperparameter}
%\vspace{-5pt}
All experiments are conducted on two pre-trained cross-lingual models. The XLM-RoBERTa-base and BERT-base Multilingual-Cased model are fine-tuned.
The number of Transformer blocks is 12, the hidden layer size is 768, the number of self-attention heads is 12, and the total number of parameters for the XLM-RoBERTa-base model is 270M, and BERT base Multilingual-Cased is 110M. 
When fine-tuning, we keep the dropout probability at 0.1 and the initial learning rate is 2e-5.
%When fine-tuning, we keep the dropout probability at 0.1 and set the number of epochs to 2 and 4. The initial learning rate is 2e-5, and the batch size is 3 and 16.
In the Preliminary Experiment (Appendix\S\ref{Section D}), %we aimed to introduce a solid regularization effect for the incoherence of the trained data by using a small batch size \cite{sekhari2021sgd}. Additionally, for fair comparison, 
we set the batch size to 3, %allowing variability in the training pattern of the input form in NLI. 
this setting was applied to both single and NLI tasks. The max length was set to 512, and for epochs beyond 3, no significant performance improvement was observed, so the results from epoch 2 were noted. Subsequently, in Main Experiment (Section\S\ref{Section 4.4}), we fine-tuned with a batch size of 16, following the pattern of previous experiments \cite{karimi-etal-2021-improving}. All models trained on Korean data achieved faster convergence and optimal performance. As no significant improvements were observed beyond 10 epochs, results were reported at 4 epochs. Each reported metric is the average of runs with four different random seeds to mitigate the effects of random variation in the results.
%\begin{comment}
%\newpage 
\begin{table}[h]
\centering\resizebox{.875\linewidth}{!}
{\begin{tabular}{lc}
\toprule
\textbf{Hyperparameter} & \textbf{Value} \\
\midrule
Model & \texttt{xlm-roberta-base, bert-base-multilingual-cased} \\
%Architecture & \texttt{AutoModelForSequenceClassification} \\
%Number of epochs & 4 \\
%Batch size (per device) & 16 \\
Learning rate & 2e-5 \\
Optimizer & AdamW \\
Weight decay & 0.01 \\
%Warmup steps & 10\% of total steps \\
Dropout (hidden layers \& attention) & 0.1 \\
Hidden size & 768 \\
Intermediate size & 3072 \\
Vocabulary size & 250K, 110K  \\
%hidden layers & 12 \\
%attention heads & 12 \\
%LayerNorm epsilon & 1e-5 \\
%Max position embeddings & 514 \\
\bottomrule
\end{tabular}}
\caption{Hyperparameters of main phase for post-training on KR3 dataset}
\label{tab:hyperparams}
\end{table}
%\end{comment}
\vspace{-5pt}
\subsection{Metrics}
%\vspace{-5pt}
The benchmarks for SemEval-2014 Task 4 are the several best performing systems in \citet{sun-etal-2019-utilizing}, \citet{wang-etal-2016-attention} and \citet{pontiki-etal-2014-semeval}.
%ATAE-LSTM \cite{wang-etal-2016-attention}. 
When evaluating Kor-SemEval and KR3 test data with subtask 3 and 4, following \citet{sun-etal-2019-utilizing}, we also use Micro-F1 and accuracy respectively. 
\vspace{-15pt}
\section{Preliminary Experiment}\label{Section D}
\subsection{Experiment: Kor-SemEval}
We conducted evaluations for each of the mBERT-single, XLM-R\textsubscript{Base}-single, mBERT-NLI, XLM-R\textsubscript{Base}-NLI, and NLI-ensemble models. 
As there is no officially converged dataset and model research specifically for Korean ABSA, We included the results from the previous SemEval14 research and Kor-SemEval to compare and evaluate the performance in Korean.

	\begin{table}[h]
          \begin{minipage}{.475\textwidth}
		\centering%\resizebox{\linewidth}{!}{
		\begin{tabular}{l  c  c  c}
		      \toprule
                %\midrule 
			\multirow{2}*{\textbf{Model}} & \multicolumn{3}{c}{\textbf{SemEval-14}}\\
			\cline{2-4}
			~ & Precision & Recall & Micro-F1 \\
                \midrule        
                BERT-single & 92.78 & 89.07 & 90.89 \\
                BERT-pair-NLI-M & 93.15 & 90.24 & 91.67 \\  
                \hline
   			%\midrule
                %\rowcolor{gray!20}
		    %\baseline{} 
                \multicolumn{4}{l}{\baseline{\emph{Models trained \& evaluated on \textbf{Kor-SemEval}}}}\\
			%\cline{2-4}  
			%~ & 4-way acc & 3-way acc & Binary  \\
			%\midrule	
			mBERT-single & 92.16 & 77.95 & 84.46 \\
			XLM-R\textsubscript{Base}-single & 91.01 & 49.37 & 64.01 \\
			mBERT-NLI & 91.10 & 79.90 & 85.14 \\
			XLM-R\textsubscript{Base}-NLI & \baseline{91.37} & \baseline{\textbf{83.71}} & \baseline{\textbf{87.37}} \\
			NLI-ensemble & \textbf{93.70} & 81.27 & 87.04 \\
                %\midrule 
                \bottomrule
		\end{tabular}
		\caption{\label{tab2} Test set results for Aspect Category Detection. We use the results reported in BERT-single and BERT-pair-NLI-M \cite{sun-etal-2019-utilizing} for English dataset together with our results.}
          \end{minipage}%
          \hspace{0.05\textwidth} % Space between the two tables
          \begin{minipage}{.475\textwidth}
		\centering%\resizebox{\linewidth}{!}{
		\begin{tabular}{l c c c}
                \toprule
               % \midrule 
			\multirow{2}*{\textbf{Model}} & \multicolumn{3}{c}{\textbf{SemEval-14}}\\
			\cline{2-4}  
			~ & 4-way acc & 3-way acc & Binary \\
			\midrule
			BERT-single & 83.7 & 86.9 & 93.3 \\		
			BERT-pair-NLI-M & 85.1 & 88.7 & 94.4 \\
   			%\midrule
                \hline 
                %\rowcolor{gray!20}
		    %\baseline{} 
                \multicolumn{4}{l}{\baseline{\emph{Models trained \& evaluated on \textbf{Kor-SemEval}}}}\\
			%\cline{2-4}  
			%~ & 4-way acc & 3-way acc & Binary  \\
			%\midrule
			mBERT-single & 68.20 & 71.84 & 79.52 \\
			XLM-R\textsubscript{Base}-single & 62.93 & 66.29 & 75.20 \\
			mBERT-NLI & 73.95 & 77.90 & 84.87 \\
			XLM-R\textsubscript{Base}-NLI & \baseline{\textbf{79.41}} & \baseline{\textbf{83.66}} & \baseline{\textbf{89.98}} \\
			NLI-ensemble & 78.24 & 82.43 & 89.65 \\
               % \midrule 
                \bottomrule
		\end{tabular}
		\caption{\label{tab3} Test set accuracy (\%) for Aspect Category Polarity. We use the results reported in BERT-single and BERT-pair-NLI-M \cite{sun-etal-2019-utilizing} for English dataset together with our results.}
          \end{minipage}
	\end{table}
\vspace{-15pt}
\subsection{Preliminary Results}
Results on Kor-SemEval are presented in Table \ref{tab2} and Table \ref{tab3}. Similar to the SemEval results, it was confirmed that tasks converted to NLI tasks tend to be better than single tasks, with mBERT achieving better results in single and XLM-R\textsubscript{Base} in NLI. The XLM-R\textsubscript{Base}-NLI model performs best, excluding precision for aspect category detection. It also works best for aspect category polarity. The NLI-ensemble model was the best in precision but performed poorly in other metrics.
\vspace{-10pt}
\section{Attention-Head Visualization}
\begin{figure*}[t]
\centerline{\includegraphics[width=\linewidth, height=15cm]{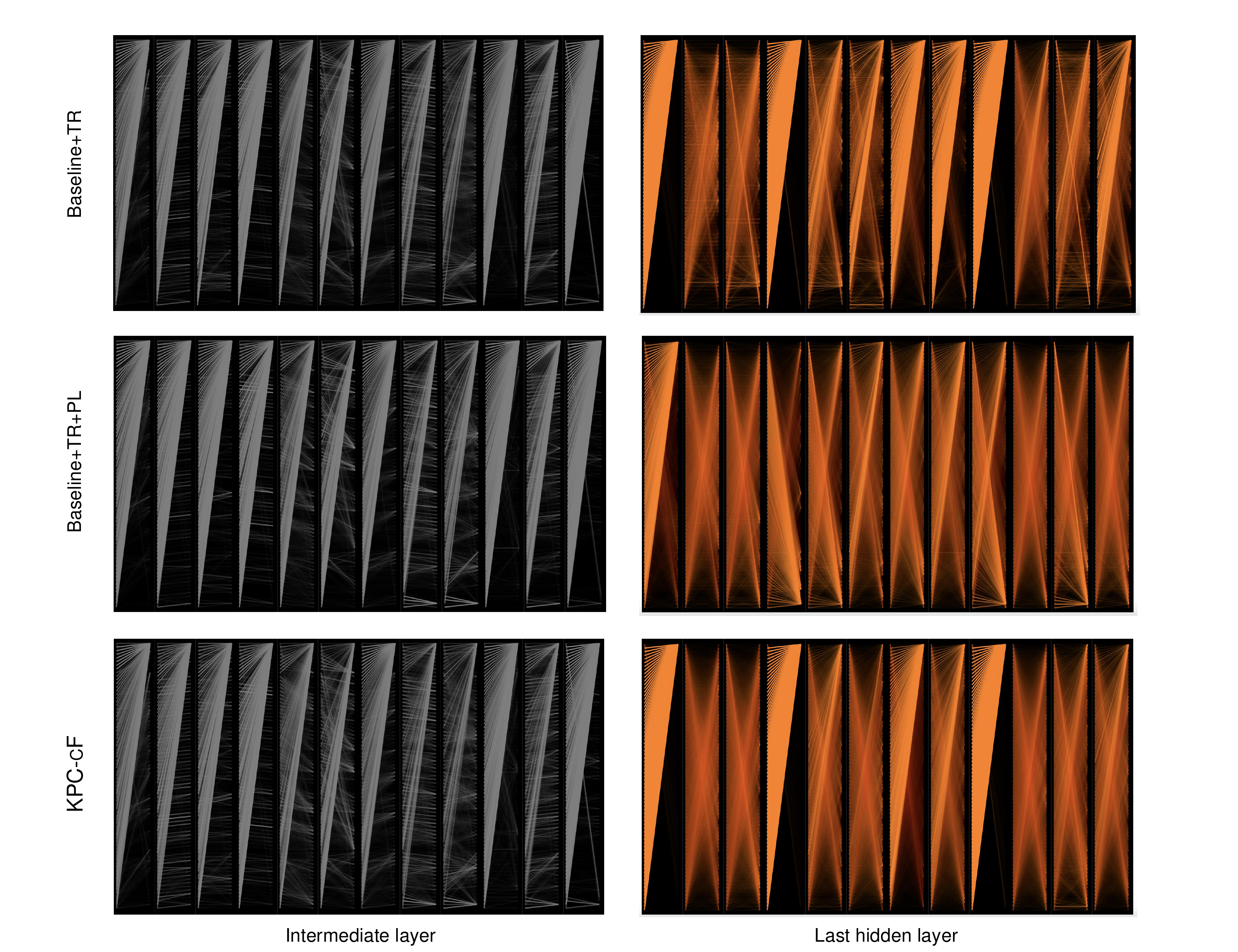}}
%\footnotesize\quad\quad \textbf{(a) TR w/o PL-\textsc{cF} \qquad\qquad\quad (b) TR w/ PL-\textsc{cF}}
\caption{
    \textbf{Multi-Head Attention Visualization.}  
    In intermediate layers associated with domain and linguistic understanding \cite{clark-etal-2019-bert}, the proposed \textsc{KPC-cF} model exhibits stronger self-attention activations across heads for tokens relevant to target language relations. In the final encoder layer responsible for classification, \textsc{KPC-cF} shows greater focus on label-relevant information in the \texttt{[CLS]} embedding, with reduced influence from irrelevant noise. This visualization is consistent with Lemma~\ref{lemma:compactness}.}
\label{fig6}
\end{figure*}

\begin{figure*}[t]
\centerline{\includegraphics[width=\linewidth, height=15cm]{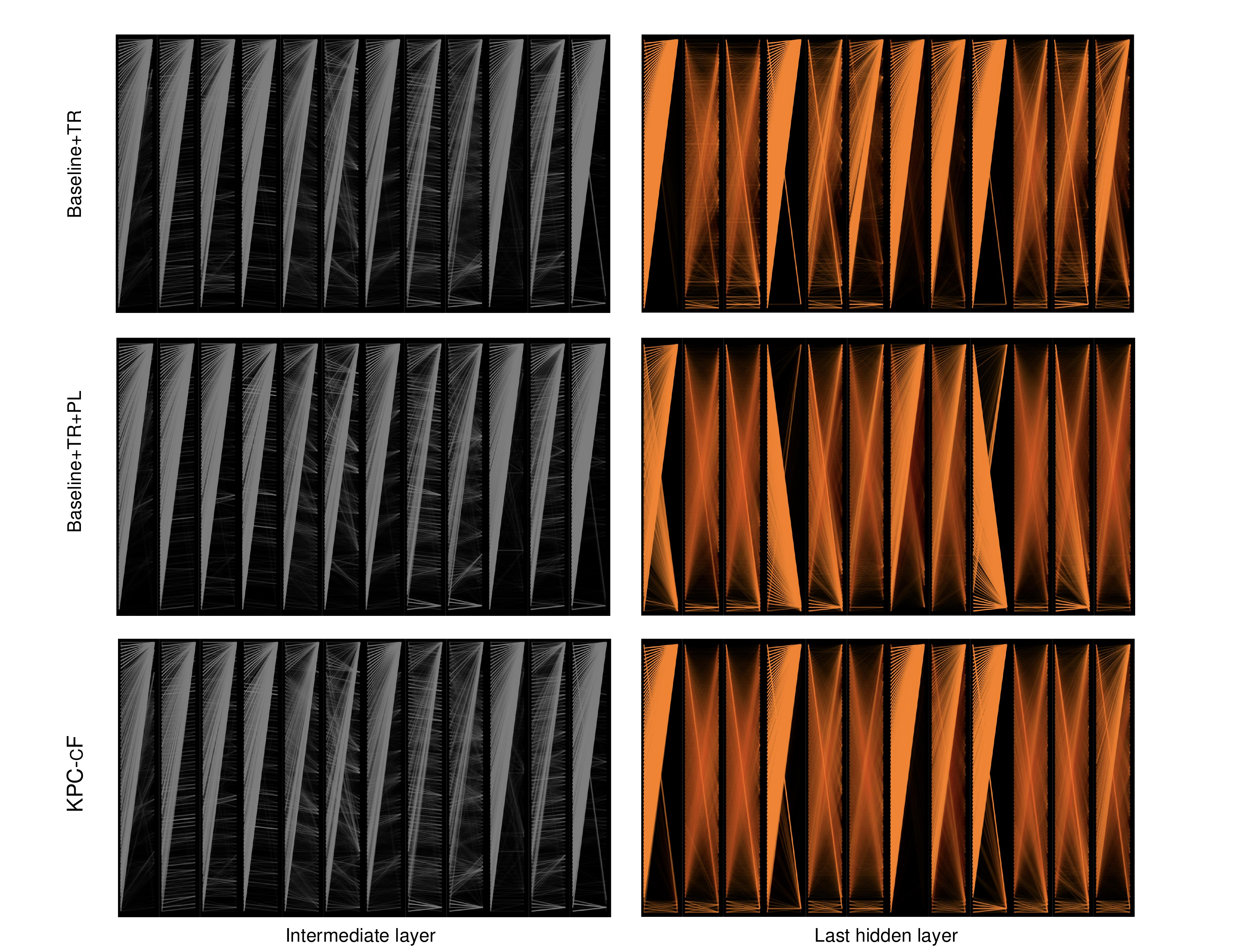}}
%\footnotesize\quad\quad \textbf{(a) TR w/o PL-\textsc{cF} \qquad\qquad\quad (b) TR w/ PL-\textsc{cF}}
\caption{
    \textbf{Multi-Head Attention Visualization.}  
    In intermediate layers associated with domain and linguistic understanding \cite{clark-etal-2019-bert}, the proposed \textsc{KPC-cF} model exhibits stronger self-attention activations across heads for tokens relevant to target language relations. In the final encoder layer responsible for classification, \textsc{KPC-cF} shows greater focus on label-relevant information in the \texttt{[CLS]} embedding, with reduced influence from irrelevant noise. This visualization is consistent with Lemma~\ref{lemma:compactness}. Snippet is as follows: review = ``맛 집이라고 찾아서 갔는데 손님이 오든 말든 불친절에 물컵에 기름기가 잘잘 흐르고 위생도 꽝이고 면에서는 밀가루 냄새만 나고 먹다가 다 남기고 실망하고 나왔어요. 왜 이런 집이 맛 집으로 평가받는지 모르겠다.'', aspect = ``서비스''.}
\label{fig7}
\end{figure*}

\begin{figure*}[t]
\centerline{\includegraphics[width=\linewidth, height=15cm]{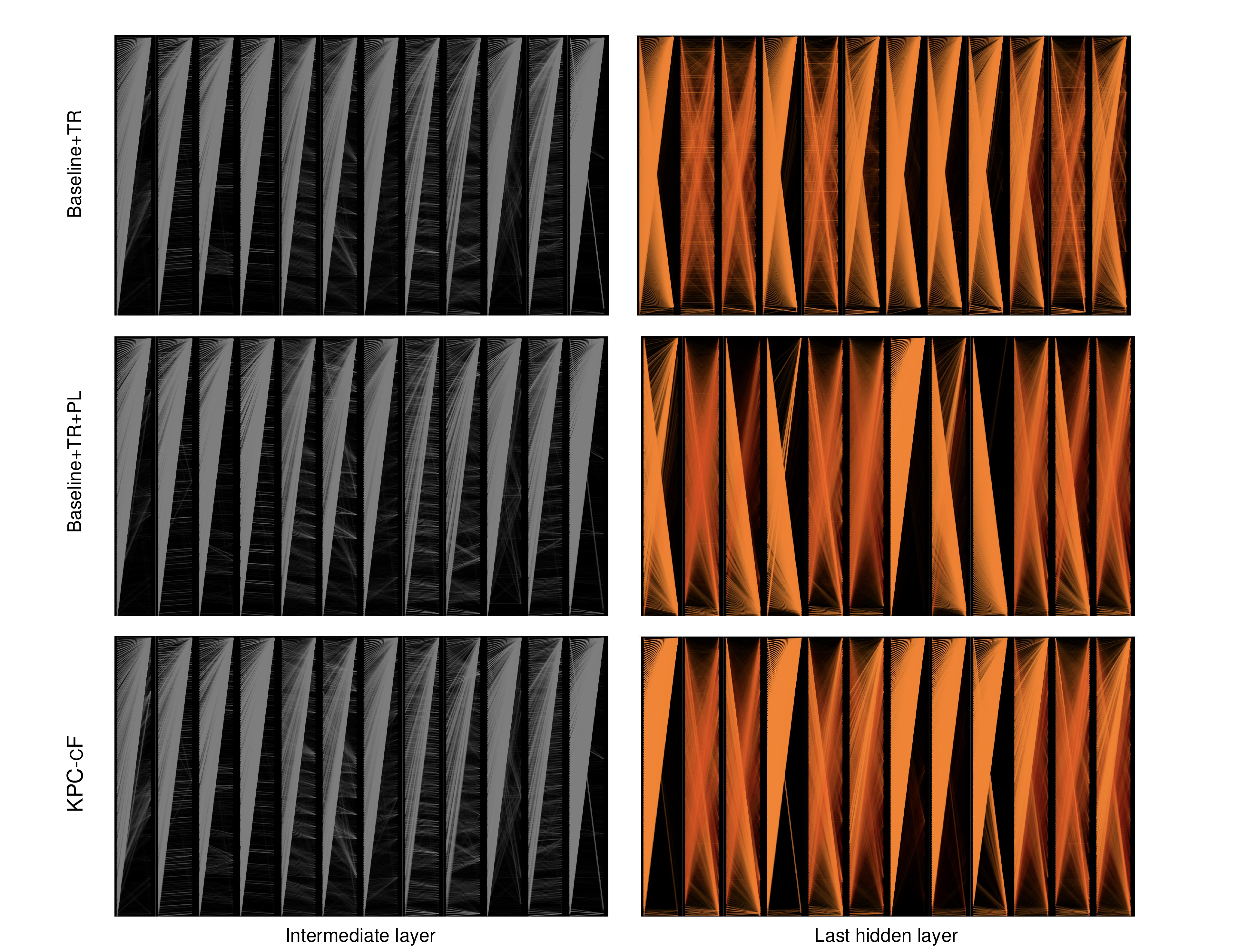}}
%\footnotesize\quad\quad \textbf{(a) TR w/o PL-\textsc{cF} \qquad\qquad\quad (b) TR w/ PL-\textsc{cF}}
\caption{
    \textbf{Multi-Head Attention Visualization.}  
    In intermediate layers associated with domain and linguistic understanding \cite{clark-etal-2019-bert}, the proposed \textsc{KPC-cF} model exhibits stronger self-attention activations across heads for tokens relevant to target language relations. In the final encoder layer responsible for classification, \textsc{KPC-cF} shows greater focus on label-relevant information in the \texttt{[CLS]} embedding, with reduced influence from irrelevant noise. This visualization is consistent with Lemma~\ref{lemma:compactness}. Snippet is as follows: review = ``편안한 느낌의 토스트 킴. 식사 한 끼로 좋은 토스트를 팝니다. 고기를 좋아해서 갈 때마다 시키는 건 불고기, 삼겹살, 돈가스, 소고기 치즈 토스트 등... 각자의 매력이 있고 한국식 토스트인데 정말 두둑합니다... 단품 세트로 시키면 감자튀김에 음료도 나오니 웬만한 패스트푸드 보다 낫다 보시면 되겠네요... 오천 원 하나로 두둑이 먹을 수 있어요.'', aspect = ``가격''.}
\label{fig8}
\end{figure*}

For token visualization (Figure \ref{fig6}, \ref{fig7}, \ref{fig8}), we used input sentences from the KR3 dataset, structured as \texttt{[CLS]} review \texttt{[SEP]} aspect \texttt{[SEP]}. An example snippet in  Figure \ref{fig6} is as follows: review = ``맛 집이라고 찾아서 갔는데 손님이 오든 말든 불친절에 물컵에 기름기가 잘잘 흐르고 위생도 꽝이고 면에서는 밀가루 냄새만 나고 먹다가 다 남기고 실망하고 나왔어요. 왜 이런 집이 맛 집으로 평가받는지 모르겠다.", aspect = ``음식".
\vspace{-10pt}
\section{Computational Resources}
All experiments presented in this study were conducted using high-performance computing resources provided by a leading cloud platform. The computational infrastructure utilized for the experiments included instances with 50GB or more of RAM and GPUs, specifically the NVIDIA A100 and V100 models. The choice of these GPUs was made to harness the advanced parallel processing capabilities essential for the computational demands of our LMs.
\vspace{-10pt}
\begin{table*}[h]
\centering{\resizebox{\linewidth}{!}{
\renewcommand{\arraystretch}{1.45}
{\LARGE\begin{tabular}{lrcccccccccccc}
\toprule
\multirow{2}*{\textbf{Model}} &  \multirow{2}*{\textbf{GPU.t}} & \multirow{2}*{\textbf{Thruput}} & \multicolumn{2}{c}{\textbf{PL Filtering}} & \multicolumn{2}{c}{\textbf{PL Alignment}} & \multicolumn{3}{c}{\textbf{Aspect Category}} & & \multicolumn{3}{c}{\textbf{Polarity}}\\
\cline{8-10}
\cline{12-14}
 & & & LaBSE\textsubscript{avg} & \MSP{}\textsubscript{0.5} & METEOR & \textsc{SemScore} & Precision & Recall & Micro-F1 & & 4-way acc & 3-way acc & Binary \\ 
\midrule
Baseline\textsubscript{XLM-R}\textsc{+TR+PL} & 41 & 30.45K & & & 48.93 & 65.85 & 91.5 & 84.23 & 87.69 & & 84.21 & 86.57 & 90.35\\
Baseline\textsubscript{XLM-R}\textsc{+TR+PL} & 31 & 22.00K & \checkmark & & 48.84 & 66.62 & \textbf{91.63} & \textbf{85.51} & \textbf{88.44} & & 83.31 & 85.66 & 89.78\\
Baseline\textsubscript{XLM-R}\textsc{+TR+PL} & 38 & 29.53K &  & \checkmark & 48.99 & 65.75 & 90.39 & 84.18 & 87.14 & & \textbf{84.50} & \textbf{86.88} & \textbf{90.86}\\
%\textbf{\textsc{KPC-cF \dag}} & 47 & 21.30K & \checkmark & \checkmark & 87.75 & 83.52 \\
\baseline{\textbf{\textsc{KPC-cF}}} & \baseline{29} & \baseline{21.30K}  & \baseline{\checkmark} & \baseline{\checkmark} & \baseline{\textbf{49.03}} & \baseline{\textbf{66.58}} & \baseline{\textbf{92.46}} & \baseline{\textbf{84.53}} & \baseline{\textbf{88.29}} & \baseline{} & \baseline{\textbf{84.34}} & \baseline{\textbf{86.72}} & \baseline{\textbf{90.89}}\\
\bottomrule
\end{tabular}}}}
\caption{\footnotesize \textbf{Ablation Study} LaBSE-based similarity thresholding improves aspect detection, while confidence-based filtering via \MSP enhances polarity classification. The proposed \textsc{KPC-cF} achieves \textbf{complementary effects} by combining both filters, yielding robust performance with \textbf{reduced data}. Throughput denotes the total number of training samples processed across both pre- and second-stage. The joint-trained {\textsc{KPC-cF} \dag} required 47 minutes of GPU time. Evaluation of PL sample quality using METEOR and \textsc{SemScore} \cite{aynetdinov2024semscore}. Dual-filtered samples in \textsc{KPC-cF} show better input-label alignment than unfiltered PL data.}
\label{Ablation}
\end{table*}

\vspace{-0.25cm}
\vspace{-0.25cm}

%\vspace{-0.25cm}

%\vspace{-0.25cm}
\begin{comment}
\clearpage
\section{Computational Resources}
All experiments presented in this study were conducted using high-performance computing resources provided by a leading cloud platform. The computational infrastructure utilized for the experiments included instances with 50GB or more of RAM and GPUs, specifically the NVIDIA A100 and V100 models. The choice of these GPUs was made to harness the advanced parallel processing capabilities essential for the computational demands of our LMs.
\end{comment}

\end{document}

%% file: std_macros.tex
\newcommand\sX{\ensuremath{\mathcal{X}}}
\newcommand\sY{\ensuremath{\mathcal{Y}}}